\documentclass[sigconf,nonacm]{acmart}

\usepackage{pgfplots}
\usepackage{pgfplotstable}
\usepackage{siunitx}
\usepackage{booktabs}
\usepackage{amsmath}
\usepackage{dblfloatfix}
\usepgfplotslibrary{statistics}
\pgfplotsset{compat=newest}
\usepgfplotslibrary{fillbetween} 
\usepgfplotslibrary{groupplots}
\usepackage{pgfgantt}

\usepackage{multirow}
\usepackage{collcell}
\usepackage{colortbl}
\usepackage{xcolor}
\usepackage{datatool}

\usepackage{balance} 

\setcopyright{none}

\usepackage{enumitem}
\usepackage{subcaption} 
\usepackage{graphicx}
\usepackage{hyperref}
\usepackage{placeins}

\usepackage[textsize=tiny, colorinlistoftodos]{todonotes}

\newcommand{\Power}{\mathcal{KW}}

\usepackage[linesnumbered,ruled,vlined,noend]{algorithm2e}

\usepackage{cleveref}

\SetCommentSty{mycommfont}
\usepackage{amsmath} 
\newcommand\nissan{Nissan}

\newcommand{\SOCR}{{{\it SOC}^{R}}}
\newcommand{\SOCI}{{{\it SOC}^{I}}} 
\newcommand{\SOC}{{{\it SOC}}} 
\newcommand{\SOCMAX}{{{\it SOC}^{max}}} 
\newcommand{\SOCMIN}{{{\it SOC}^{min}}} 
\newcommand{\PrdPeak}{\hat{P}^{max}} 

\newcommand{\PowerNeed}{{\it KWH^{R}}} 
\newcommand{\ReTime}{\tau^R}
\newcommand{\MaskAction}{A'}

\newcommand{\Building}{\mathcal{B}}
\newcommand{\DepartureTime}{\mathcal{D}} 
\newcommand{\policyGuidanceRate}{{R^{PG}}}
\newcommand{\Mask}{{\it Mask}}
\newcommand{\MILP}{{\it MILP}} 
\newcommand{\Buffer}{\mathbf{BF}} 

\newcommand{\rlcluster}{{\bf RL\textbackslash{}500} }  
\newcommand{\rlrandom}{{\bf RL\textbackslash{}C} } 
\newcommand{\rlmorefeature}{{\bf RL\textbackslash{}F} } 
\newcommand{\random}{{\bf Random\textbackslash{}A} } 

\newcommand{\rlnop}{{\bf RL\textbackslash{}P}}
\newcommand{\rlnoa}{{\bf RL\textbackslash{}A}}

\newcommand{\rlnoe}{{\bf RL\textbackslash{}E}}

\acmSubmissionID{966}

\title[AAMAS-2025 Formatting Instructions]{Reinforcement Learning-based Approach for Vehicle-to-Building Charging with Heterogeneous Agents and Long Term Rewards }

\author{Fangqi Liu}
\affiliation{
  \institution{Vanderbilt University}
  \city{Nashville, TN}
  \country{USA}}
\email{fangqi.liu@vanderbilt.edu}

\author{Rishav Sen}
\affiliation{
  \institution{Vanderbilt University}
  \city{Nashville, TN}
  \country{USA}}
\email{rishav.sen@vanderbilt.edu}

\author{Jose Paolo Talusan}
\affiliation{
  \institution{Vanderbilt University}
  \city{Nashville, TN}
  \country{USA}}
\email{jose.paolo.talusan@vanderbilt.edu}

\author{Ava Pettet}
\affiliation{
  \institution{Nissan Advanced Technology Center - Silicon Valley}
  \city{Santa Clara, CA}
  \country{USA}}
\email{ava.pettet@nissan-usa.com}

\author{Aaron Kandel}
\affiliation{
  \institution{Nissan Advanced Technology Center - Silicon Valley}
  \city{Santa Clara, CA}
  \country{USA}}
\email{aaron.kandel@nissan-usa.com}

\author{Yoshinori Suzue}
\affiliation{
  \institution{Nissan Advanced Technology Center - Silicon Valley}
  \city{Santa Clara, CA}
  \country{USA}}
\email{yoshinori.suzue@nissan-usa.com}

\author{Ayan Mukhopadhyay}
\affiliation{
  \institution{Vanderbilt University}
  \city{Nashville, TN}
  \country{USA}}
\email{ayan.mukhopadhyay@vanderbilt.edu	}

\author{Abhishek Dubey}
\affiliation{
  \institution{Vanderbilt University}
  \city{Nashville, TN}
  \country{USA}}
\email{abhishek.dubey@vanderbilt.edu}

\begin{abstract}
{\color{black} 
Strategic aggregation of electric vehicle batteries as energy reservoirs can optimize power grid demand, benefiting smart and connected communities, especially large office buildings that offer workplace charging. 
This involves optimizing charging and discharging to reduce peak energy costs and net peak demand, monitored over extended periods (e.g., a month), which involves making sequential decisions under uncertainty and delayed and sparse rewards, a continuous action space, and the complexity of ensuring generalization across diverse conditions. Existing 
algorithmic approaches, e.g., heuristic-based strategies, fall short in addressing real-time decision-making under dynamic conditions, and traditional reinforcement learning (RL) models struggle with large state-action spaces, multi-agent settings, and the need for long-term reward optimization. To address these challenges, we introduce a novel RL framework that combines the Deep Deterministic Policy Gradient approach (DDPG) with action masking and efficient MILP-driven policy guidance. Our approach balances the exploration of continuous action spaces to meet user charging demands. 
Using real-world data from a major electric vehicle manufacturer, we show that our approach comprehensively outperforms many well-established baselines and several scalable heuristic approaches, 
achieving significant cost savings while meeting all charging requirements. Our results show that the proposed approach is one of the first scalable and general approaches to solving the V2B energy management challenge.
}

\end{abstract}



\newcommand{\BibTeX}{\rm B\kern-.05em{\sc i\kern-.025em b}\kern-.08em\TeX}

\begin{document}

\pagestyle{fancy}
\fancyhead{}

\maketitle 

\section{Introduction}
\label{section:introduction}

The concept of vehicle-to-building (V2B)  charging \cite{kempton2005vehicle, lund2008integration} leverages the ability of battery electric vehicles (EVs) to operate as both energy consumers and temporary storage units \cite{tomic2007using}. V2B systems are particularly relevant in large office buildings, where EVs can be aggregated to optimize energy consumption and reduce peak power demand. By strategically controlling the charging and discharging cycles of EVs, these systems ensure that vehicles meet users' expected state-of-charge (SoC) requirements while minimizing the energy bought during peak time-of-use (ToU) periods  \cite{tse2014use, zhao2023research} and reducing the building's peak power demand over a billing cycle.
Implementing this optimization process in practice becomes complex due to the heterogeneity of charging infrastructures \cite{park2024exact}, the uncertainty of EV arrival and departure times, and the need for a careful balance between energy cost savings and ensuring that the expected final state of charge (SoC) is kept close to user expectation.  Additionally, aligning V2B frameworks with complex electricity pricing policies, including both energy and demand charges, adds to the challenge \cite{zhang2018optimal, 8274175}. While prior work has largely modeled this problem as a single-shot mixed-integer linear program~\cite{AORC2013, 5986769, 9409126, MJG2015}, such approaches fail to capture the intricacies of real-time decision-making in dynamic environments.

This sequential decision process can be modeled as a Markov Decision Process (MDP); however, solving the MDP presents several difficulties, including delayed and sparse rewards, a continuous action space, and the need for effective long-term decision-making under uncertainty. 
To address these challenges, we propose a novel approach to solve this problem that combines the Deep Deterministic Policy Gradient (DDPG) with two key enhancements: action masking and policy guidance through a mixed-integer linear program (MILP). The DDPG algorithm allows us to optimize continuous action spaces while accounting for uncertainties in EV arrival times, SoC requirements, and fluctuating building energy demands. By leveraging action masking, we adjust neural network actions during training using domain-specific knowledge, limiting exploration and guiding the RL agent toward more efficient and feasible policies.  The MILP component provides policy guidance during training, steering the RL agent toward near-optimal solutions and enhancing convergence in complex environments. 
Our approach demonstrates strong generalization across diverse conditions and offers a scalable solution for V2B energy management.
Our team includes a major EV manufacturer with access to a smart building that has 15 heterogeneous chargers (~\Cref{fig: EV chargers} shows some of them). We use real-world charging and energy data to validate our approach, showing its effectiveness in reducing energy costs over nine months (May 2023 – Jan 2024). The summary of our contributions is as follows:  
\begin{figure}[t]
    \centering
    \includegraphics[width=0.58\linewidth]{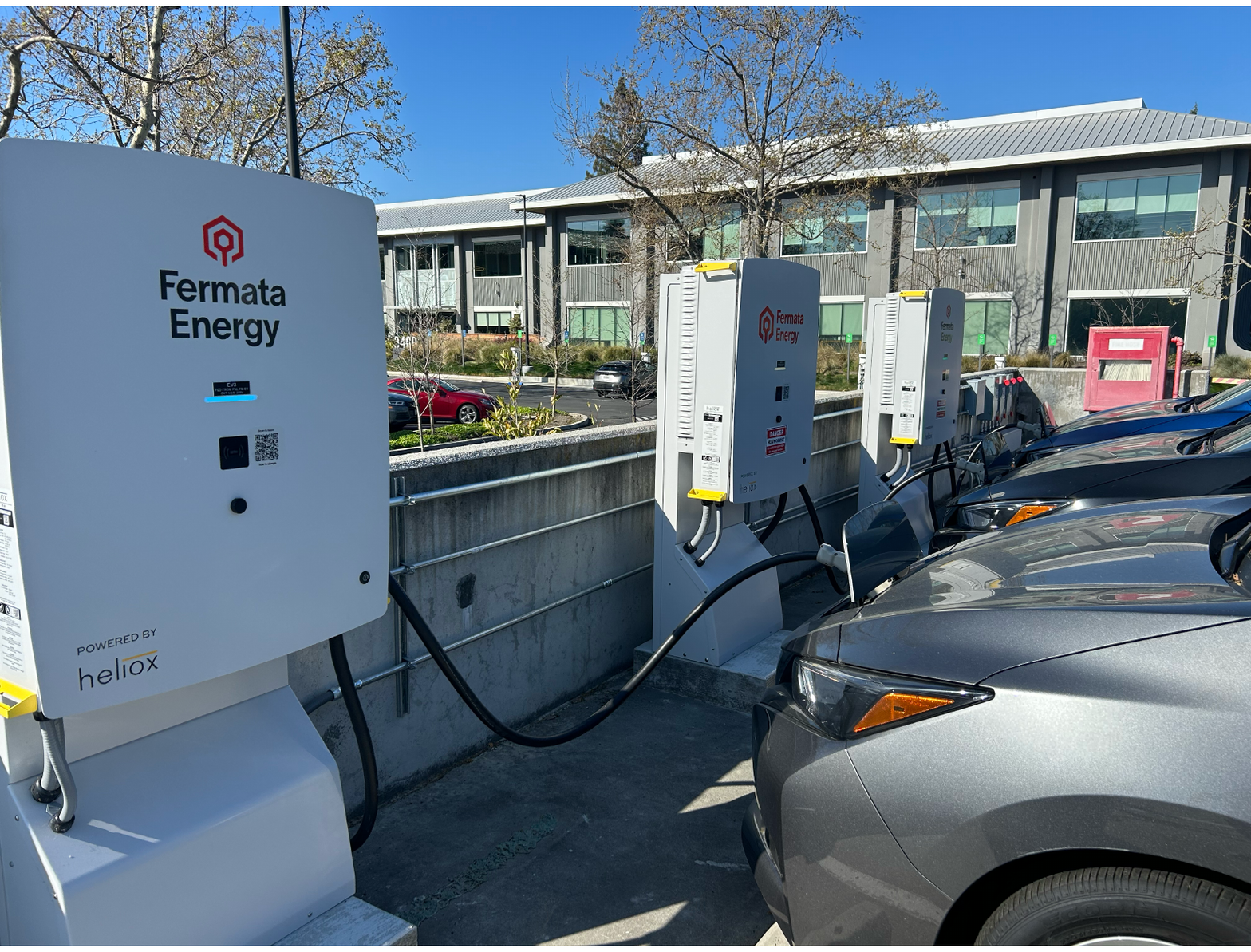}
    \caption{EVs and bidirectional chargers at the research site.}
    \label{fig: EV chargers}    
\end{figure}  
\begin{itemize}[leftmargin=*]
\item \textbf{Modeling the V2B problem as an MDP with continuous action space}: We model the V2B problem as a Markov Decision Process (MDP) that captures the dynamics of EV SoC levels, varying arrival and departure times, and time-dependent electricity pricing. This formulation addresses delayed and sparse rewards, continuous action spaces, and long-term goals to reduce the monthly peak demand charge and energy costs. 
\item \textbf{Solving the V2B sequential decision-making problem}: We present a novel RL framework based on the Deep Deterministic Policy Gradient (DDPG). We combine DDPG with i) action masking that leverages domain knowledge and the structure of the V2B problem and ii) policy guidance based on solving a deterministic MILP to aid the learning of the optimal policy.

\item \textbf{Validating with real-world data}: We validate our proposed approach using real-world data from a major electric vehicle manufacturer. The model achieved significant cost savings
over nine months (May 2023–January 2024), meeting all user charging demands. Our approach outperforms heuristics and prior work.
\item \textbf{Ablation Study:} We conduct a detailed ablation study to assess the impact of each technique and demonstrate the model's effectiveness.



\end{itemize} 


\section{Problem Formulation}
\label{sec:problem_statement}

\noindent \textbf{Charger and Time Intervals}: Consider the building has $N$ heterogeneous chargers $\mathcal{C} = \{C_1, C_2, \dots,C_N\}$. Each charger $C_i$ has limits on the charging rate, minimum $C_i^{min}$ and maximum $C_i^{max}$; $C_i^{min} < 0$ implies the charger $C_i$ is bi-directional and can discharge and $C_i^{min} = 0$ represents a unidirectional charger with no discharging. We assume that all chargers are designed to be able to charge at maximum rates simultaneously, i.e., $\sum_{i=1}^{i=N} C_i^{max} < \text{maximum rated capacity of the building} $.  
The planning horizon is one billing period, usually a month, which we divide into equal-sized fixed time intervals $\mathcal{T} = \{T_1, T_2, \dots\, T_{end}\}$, where $T_{j}-T_{j-1}=\delta$ (we use  $\delta$ = 0.25 hours). The choice of $\delta$ is user-specific and provides a stable decision epoch, preventing rapid changes in the charging rate.

\noindent \textbf{Charging Power}: Let us assume that the function $\mathcal{P}:  \mathcal{C} \times \mathcal{T}  \rightarrow \Re$ specifies the power consumed by the charger $C_i$ at time $T_j$. If the power is zero, the charger is not active, and if the power is negative, the charger discharges, acting as an energy source. Note that by construction $P(C_i,T_j) \in [C_i^{min},C_i^{max}]$. Let us also assume that function $\mathcal{B}: \mathcal{T}  \rightarrow \Re^{+} $ specifies the average building power consumed in $\delta$ time interval. 
Given the charger and the building power consumption, we can calculate the total cost for the billing period. The parts of the total cost are based on the property type, time of day, and state of the power grid and are based upon the rules and regulations set by the local transmission system operator (TSO) and distribution system operator (DSO). These parts include energy expenses for building power and charging, which vary with peak and off-peak hours, as well as demand charges based on the peak power draw over a longer-term period. 

Let the price of the energy consumed is given by $\theta_E : \mathcal{T}  \rightarrow \Re^{+}$ (in \$/kWh). In practice, the Time-of-Use (TOU) electricity rates do not vary continuously and are rather divided into two parts each day, i.e., a peak and a non-peak period. 
Then, the total cost of the energy consumed is  $\Theta_E(\mathcal{P})= \sum_{j=1}^{j=end} \left(\sum_{i=1}^{i=N} (P(C_i,T_j)) + \Building (T_j)\right) \times \theta_E  (T_j) \times \delta$. Effectively, $\Theta_E$ is a function of charging power  $\mathcal{P}=\{P(C_i, T_j)| C_i\in \mathcal{C}, T_j\in \mathcal{T}\}$.  
 
\noindent \textbf{Demand Charge}: The demand charge is calculated using the maximum (peak) power consumed during any time interval in the billing period, with the demand price denoted as $\theta_D$ (in \$/kW).
Let $P^{max} = \max_{j=1}^{j=end} (\sum_{i=1}^{i=N}$ $P(C_i,T_j)) + \Building(T_j)$ denote the maximum power consumed. The demand charge is given by $\Theta_D(\mathcal{P})= \theta_D \times P^{max} \times \delta$, which is a function of charging power  $\mathcal{P}$. Hence, the total cost of energy bought from the power grid is  $\Theta_E(\mathcal{P})+\Theta_D(\mathcal{P})$. To minimize the cost, we must reduce the net power usage when the cost $\theta_E$ is high and manage the power peaks to ensure $P^{max}$ remains as low as possible. Often, the demand charge is levied to ensure that the industrial buildings do not put excess burden on the power grid. In our problem, we use estimates of peak power and denote it by $\hat{P}^{max}$. It is important to note that the demand charge is typically applied during peak hours of the TOU electricity rate, as reflected in our formulation.

\noindent \textbf{Electric Vehicle Sessions}: Assume that during the billing period $\mathcal{T}$, a set of electric vehicles, denoted as $\mathcal{V}$, are serviced at the building. Each EV $V$ is characterized by its arrival time $\mathcal{A}: \mathcal{V} \rightarrow  \mathcal{T}$ and departure time $\mathcal{D}: \mathcal{V} \rightarrow  \mathcal{T}$. Note that if the same vehicle arrives more than once, we will treat it as a separate session. If the EV arrives between time slots $[T_{i-1}, T_{i}]$, we consider its effective arrival time as $\mathcal{A}(V) = T_i$. Similarly, if the vehicle departs between $[T_{j}, T_{j+1}]$, we consider its effective departure time as $\mathcal{D}(V) = T_j$. EV sessions are contiguous, i.e., EV is expected to remain at the site between $\mathcal{A}(V)$ and $\mathcal{D}(V)$, for $\forall V \in \mathcal{V}$. 
For each  $V$, we know the initial state of charge   $\SOCI: \mathcal{V} \rightarrow \Re^+$ and the required final state of charge (measured as a percentage of the battery capacity)   $\SOCR: \mathcal{V} \rightarrow \Re^+$ upon arrival. $\SOCMIN: \mathcal{V} \rightarrow \Re^+$ is the minimum allowed SoC for the car i.e., the car cannot be discharged below this value, and $\SOCMAX: \mathcal{V} \rightarrow \Re^+$  is the maximum allowed SoC for the car. The minimum and maximum bounds are specified by the EV manufacturer, considering the impact of charging and discharging on battery health. ${\it CAP}:\mathcal{V} \rightarrow \Re^+$ denotes the vehicle's battery capacity in kWh. We track the current SoC of the EV using ${\it SOC}$, where ${\it SOC}: \mathcal{V} \times \mathcal{T} \rightarrow \Re^+$ and it is defined later.

\noindent \textbf{Charger Assignment}: 
{
Our approach employs a two-layer decision-making process for EV charging optimization. First, a heuristic assigns EVs to chargers upon arrival. Second, an RL-based policy optimizes charging rates at fixed intervals. 
}
We define an EV assignment function $\eta: \mathcal{V} \rightarrow \mathcal{C}$, where ($V \in \mathcal{V}$) $\eta(V) = C_i$ indicates the charger assigned to EV $V$. Correspondingly, we also maintain a charger-EV occupancy function $\phi: \mathcal{C} \times \mathcal{T} \rightarrow \mathcal{V}$, where $\phi(C_i, T_j) = V$, representing the connection of charger $C_i$ with EV $V$ at time $T_j$.  
The correlation of these two functions can be expressed as $\phi(\eta(V), T_j) = V, \ \text{s.t.}\ \mathcal{A}(V) \leq T_j \leq \mathcal{D}(V) $
indicating that if EV $V$ is assigned to charger $C_i$ through the function $\eta$, then at any time slot within its stay duration, it is confirmed that EV $V$ is connected to charger $C_i$. If no EV is connected to the charger at time $T_j$, the function may return a $\emptyset$ denoting an inactive state, expressed as $\phi(C_i, T_j) = \emptyset$. 
This underscores the dynamic nature of charger assignments, which ensures that no two electric vehicles share a charger simultaneously. Our FIFO policy prioritizes bidirectional chargers as the optimal strategy (see ~\Cref{table:charger_assignment_policies} in the appendix\footnote{The full paper, including the appendix, is available on arXiv.}), enhancing charging efficiency. 
We also maintain the connection between the assigned charger and the EV until departure. For EV charging, we approximate a linear charging profile, following prior work~\cite{sundstrom2010optimization}. The SoC is updated at each time slot  $T_j$  using the following equation: 
\begin{equation}
{\it SOC}(V, T_{j+1}) = {\it SOC}(V, T_j) + \textstyle\frac{P(\eta(V), T_j)\times \delta} {{\it CAP}(V)}
\label{eq: soc}
\end{equation}

\noindent \textbf{Feasibility}:
The set \textit{Feasible} indicates the feasible solutions that satisfy the following constraints:
\begin{align}
    & \forall C_i \in \mathcal{C}, \forall T_j \in \mathcal{T}: C_i^{min} \leq P(C_i, T_j) \leq C_i^{max} \label{eq:charging_rate} \\
    & \forall C_i \in \mathcal{C}, \forall T_j \in \mathcal{T}, \forall V \in \mathcal{V}: {\it SOC}(V, T_j) \geq \SOCMIN(V)\label{eq:soc_min} \\
    & \forall C_i \in \mathcal{C}, \forall T_j \in \mathcal{T}, \forall V \in \mathcal{V}: {\it SOC}(V, T_j)\leq \SOCMAX(V)\label{eq:soc_max} \\
    & \forall T_j \in \mathcal{T}: \textstyle\sum_{C_i \in \mathcal{C}} P(C_i, T_j) + \mathcal{B}(T_j) \geq 0  \label{eq:building_power}     
\end{align} 
Here, Constraint~(\ref{eq:charging_rate}) guarantees a valid charging action range, Constraints~(\ref{eq:soc_min} and \ref{eq:soc_max}) ensures that each EV's SoC remains within an acceptable range, and Constraint~(\ref{eq:building_power}) ensures that discharging power does not exceed building power.

\noindent \textbf{Objectives}: 
One of our objectives for the V2B problem is to minimize the total cost over the billing period, incorporating the Time-Of-Use (TOU) electricity rates and demand charges. This objective is expressed as:
\begin{align}
\label{eq: billing}
\begin{split}
\min_{(\eta,\mathcal{P}) \, \in \textit{Feasible}}
\left( \Theta_E (\mathcal{P}) + \Theta_D(\mathcal{P}) \right)
\end{split}
\end{align}

The second objective ensures that vehicles are charged to their requirement, $\SOCR$, by the time they leave.
\begin{align}
\label{eq: soc_penalty}
\begin{split}
\min_{(\eta,\mathcal{P}) \in \textit{Feasible}} \textstyle\sum_{V \in \mathcal{V}} \max(\SOCR(V) - {\it SOC}(V,\mathcal{D}(V)), 0)
\end{split}
\end{align}
The inner \texttt{max} function ensures EV users' energy requirements are met, even if overcharging occurs.
However, in practical scenarios, short stays may make meeting the SoC requirement impossible. To address this, we reformulate the objectives into a multi-weighted framework.
The optimal charger assignment and actions are then determined by optimizing these combined objectives.

\section{Related Work}
\label{sec:related_work}

We highlight four major challenges of solving the V2B problem, namely: 1) the uncertainty of vehicles and SoC requirements; 2) Time-Of-Use (TOU) pricing, demand charges, and long-term rewards; 3) heterogeneous chargers and continuous action spaces; and 4) tracking real-world states and transitions. Below, we briefly cover prior work to tackle these challenges. \textit{A more detailed description of prior work is presented in~\Cref{tab:comparison} of the appendix.}  

\noindent\textbf{Uncertainty of vehicles and SoC requirements.} 
Meta-heuristics and Model Predictive Control (MPC) have been used to solve the EV charging process, focusing on energy cost and user fairness in single-site or vehicle-to-grid (V2G) systems~\cite{AORC2013, 5986769, 9409126, MJG2015}. 
Studies by \citeauthor{richardson2011electric} analyze EV charging strategies' impact on grid stability, relevant to V2B systems~\cite{richardson2011electric}. \citeauthor{8274175} proposed a demand response framework for optimizing V2B systems amidst dynamic energy pricing~\cite{8274175}. Additionally, \citeauthor{oconnell2010integration} utilized Mixed Integer Linear Programming (MILP) to integrate renewable energy sources into grids~\cite{oconnell2010integration}.
However, many of these methods focus on unidirectional chargers and fail to fully account for all exogenous sources of uncertainty (e.g., uncertain arrival and departure times).


\noindent\textbf{Time of use pricing, demand charge, and long-term rewards.} 
V2B optimization is difficult due to long billing periods. While prior work (barring some exceptions~\cite{9409126}) optimizes and plans for single-day horizons~\cite{AORC2013, MMN2019, SNDJ2020}, they fail to work for longer periods.

\noindent\textbf{Heterogeneous chargers and continuous action spaces.} 
In practice, buildings develop EV infrastructure gradually, leading to heterogeneous chargers and a more complex action space.
While some prior work addresses charger heterogeneity~\cite{NNM2024,ZJS2022}, it often neglects long-term rewards (i.e., limit planning to a single day) or fails to account for demand charge, missing the key real-world constraint in the V2B problem.
\noindent\textbf{Tracking real-world state and transition.}
Existing solutions validate their approaches using simulations with limited interface with the real world (barring some exceptions~\cite{9409126}), thereby making simplistic assumptions that limit deployment.

\section{Our Approach}
\label{sec:our_approach}
\begin{figure*}[t]
    \centering
    \begin{subfigure}[b]{0.68\linewidth}
        \includegraphics[height=2.1in,keepaspectratio,trim=0.2cm 0.2cm 0.2cm 0.2cm,clip]{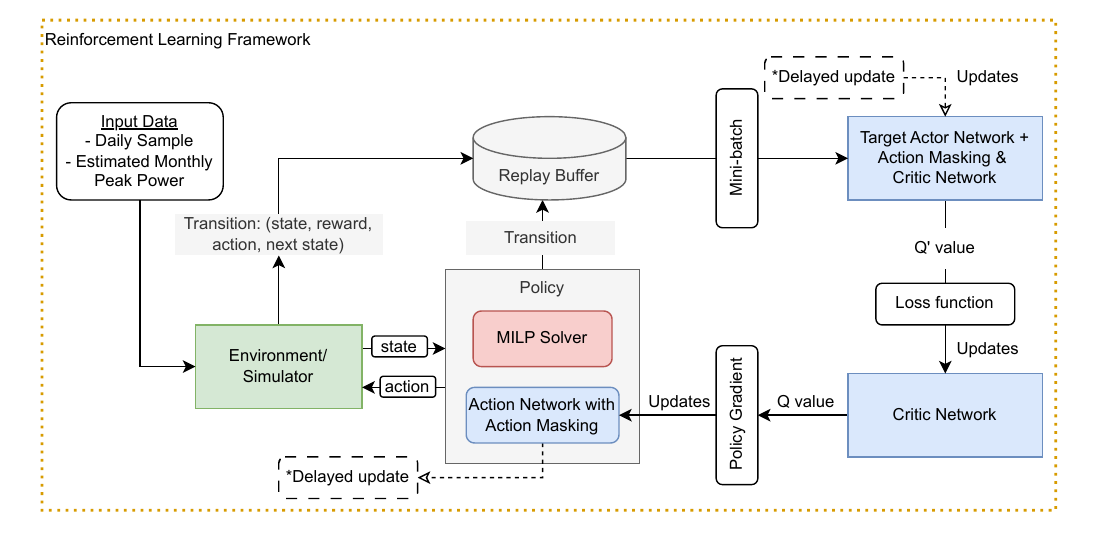}
        \caption{Reinforcement Learning Framework.}
       \label{fig:framework}
    \end{subfigure}
    \begin{subfigure}[b]{0.3\linewidth}
        \includegraphics[height=2.1in,keepaspectratio,trim=0cm 0.5cm 0cm 0.5cm,clip]{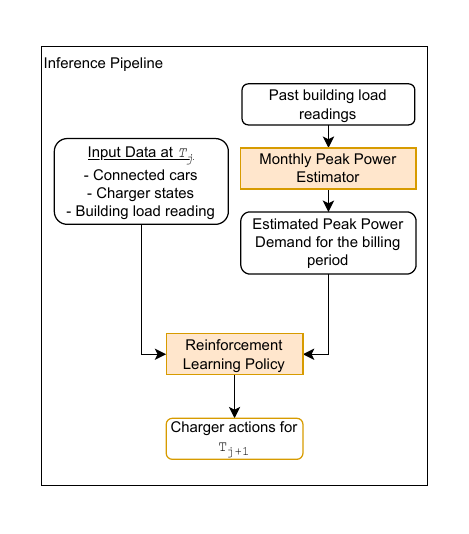}
        \caption{Pipeline for Inference.}
        \label{fig:pipeline}
    \end{subfigure}
    \caption{\color{black}(a) Our framework relies on daily samples and an estimated monthly peak power. We use RL, i.e., DDPG, and extend it with policy guidance and action masking, to learn a near-optimal policy. (b) At inference time, the model ingests data of connected cars, charger states, building power, and the estimated monthly peak power to make decisions.}
    \label{fig:framework_and_pipeline}
\end{figure*}

In this section, we discuss the different components in our  framework, shown in~\Cref{fig:framework}. 

\subsection{Markov Decision Process Model}
\label{ssec:MDP}

We model the V2B problem as the following MDP. 

{\bf State.}
The complete state space for the problem can be described using features that capture historical, current, and future estimation at a given time $T_j$, which includes parameters for each vehicle, such as the current SoC, required SoC, departure time, and battery capacity for each EV, along with SoC boundaries across all chargers. Additionally, the current building power, time slot, day of the week, historical building power, and long-term peak power estimation value are included, resulting in approximately $100$ features. 
We leverage domain-specific knowledge to abstract key information from these features, reducing the state space to the $37$ essential state elements.

These features are:
\textbf{1)} The current time slot, $T_j$. \textbf{2)} The current building power, denoted as ${B}(T_j)$. \textbf{3)} The power gap between the current building power and the estimated peak power for the billing period, given by $ \PrdPeak(T_j) - B(T_j)$, where $\PrdPeak(T_j)$ indicates the estimated peak power at $T_j$, initialized from a value derived from training data. This gap aids the RL model in estimating the optimal peak power for demand charge reduction. \textbf{4)} The mean peak building power over the previous 7 days, $\mu(B^H(T_j))$, where $B^H(T_j)$ represents the list of peak building power for the previous 7 days. \textbf{5)} {The variance of the peak building power over the previous 7 days, $\sigma^2(B^H(T_j))$, helps inform the model about the future building power use}. \textbf{6)} The day of the week for the current time slot, $T_j$, which helps the model distinguish daily patterns and enhance generalization. \textbf{7)} The number of EV arrivals up to time slot $T_j$, represented as $|\{V | V \in \mathcal{V}, A(V) \leq T_j \}|$ for tracking EV arrival status. \textbf{8)} The energy needed by each EV connected to a charger at time slot $T_j$, given by $[\PowerNeed(C_i, T_j)]_{C_i \in \mathcal{C}}$, which is initialized to $0$. This quantity represents the energy gap between required SoC ($\SOCR$) and current SoC ($\it{SOC}$) of the EV $V = \phi(C_i, T_j)$, defined as $\PowerNeed(C_i, T_j) = (\SOCR(V) - \SOC(V, T_j)) \times \text{CAP}(V)$.
\textbf{9)} The remaining time until the departure of each EV connected to the chargers is given by $[\ReTime(C_i, T_j)]_{C_i \in \mathcal{C}}$, and is set to 0 when no cars are connected. Each term is computed as $\ReTime(C_i, T_j) = \DepartureTime(\phi(C_i, T_j)) - T_j$.

{\bf Actions.} We define the set of actions $\mathcal{A}$, which includes all actions at each time slot $T_j$ with $T_j \in \mathcal{T}$. In this MDP, $\mathcal{A}$ is continuous and specifies the power of all chargers at each time slot $T_j$, where $A(T_j) = [P(C_i, T_j)]_{C_i \in \mathcal{C}} $.

{\bf State Transition.} 
States are updated based on actions and EV arrivals/departures at each time slot. To simulate these transitions, we designed an environment simulator that provides and updates states. The state transition function is given as:
${\it Trans}(S(T_{j-1})$, $A(T_{j-1})) \mapsto S(T_j)$, with the following steps: 
\begin{enumerate}[leftmargin=*]
    \item Initialize the estimated peak power, $\PrdPeak(T_0)$, which can be derived from historical data 
    (detailed in ~\Cref{sec:our_approach})
    , and update it by
    $
    \PrdPeak(T_{j}) = \max(\PrdPeak(T_{j-1})$, $ \Building(T_{j-1}) + \sum_{C_i \in \mathcal{C}} P(C_i, T_{j-1})),
    $
    which updates the estimated peak power depending on the previous estimate and the last peak power.
    \item Update SoC of EVs connected to all chargers: $\it{SOC}(\phi(C_i,T_{j}), T_{j})$ using action $A(T_{j-1})$ according to Equation~(\ref{eq: soc}). 
   \item Update the EV charger assignment $\phi(C_i, T_j)$ and $\eta(V)$ by first releasing chargers with departing EVs in the current time slot $T_j$ and then assigning new arrival EVs to idle chargers. 
   \item Update the energy requirement of all EVs connected to a charger: $[\PowerNeed(C_i, T_j)]_{C_i \in \mathcal{C}}$ by based on EV's current SoCs.
   \item Update the remaining time of all EVs connected to chargers: $[\ReTime(C_i, T_{j})]_{C_i \in \mathcal{C}}$ at time slot $T_{j}$.   
\end{enumerate}

{\bf Reward.} We define the function ${\it Reward}: \mathcal{S} \times \mathcal{A} \rightarrow \Re$, where ${\it Reward}(S(T_j), A(T_j))$ evaluates the reward for actions taken in a specific state, focusing on minimizing the total bill while satisfying SoC requirements. We express reward as $\lambda_{S} \cdot \mathit{r}_1 + \lambda_{E} \cdot \mathit{r}_2 + \lambda_{D} \cdot \mathit{r}_3$ where $\mathit{r}_1 =  \sum_{C_i\in\mathcal{C}} \max(0, \min(\PowerNeed(C_i, T_j), P(C_i, T_j) \times \delta))$, $\mathit{r}_2 = - P(C_i, T_j) \cdot \delta  \cdot \theta_E(T_j)$, and $\mathit{r}_3 = - \max(0, \Building(T_j) + \sum_{C_i \in \mathcal{C}} P(C_i, T_j) - \PrdPeak(T_j)) \cdot \theta_D$
.
In this reward structure, $\mathit{r}_1$ promotes actions that charge EVs to reach their required SoC, as intended in Equation~(\ref{eq: soc_penalty}), while $\mathit{r}_2$ penalizes the energy cost for the charging actions taken. The third component, $\mathit{r}_3$, penalizes the increase in demand charges if peak power increases, aligning with our objective in Eq. (\ref{eq: billing}). These functions use three  coefficients, $\lambda_{S}$, $\lambda_{E}$, and $\lambda_{D}$ to balance trade-offs.

\subsection{Reinforcement Learning Approach}
\label{sec:RL}
In this section, we describe the entire reinforcement learning pipeline. We introduce the network structure, discuss how we use a simulator to gather state features and describe the different techniques, such as action masking and policy guidance, used to improve the performance of the V2B problem.  

To improve training efficiency, we address the challenge of long state-action sequences by splitting the monthly dataset into daily episodes. This allows the model to capture variations across different weekdays and learn more effectively from shorter episodes, adapting more quickly to daily changes. By incorporating estimated monthly peak power into the state features and reward function, the approach still accounts for monthly demand charges, helping to minimize long-term costs while staying aligned with our objective. 

%
\subsubsection{Enhanced Deep Deterministic Policy Gradient}
Our approach based on the DDPG framework~\cite{lillicrap2015continuous} uses an actor network for continuous actions.
During training, we interact with the simulator that provides state abstractions and transitions.
To improve RL performance in handling the limitations associated with large continuous action spaces and long-term reward optimization, we introduce action masking and policy guidance techniques. Details of the enhanced approach are in Algorithm~\ref{alg:DDPG} in the appendix. 
Action masking, denoted as $\Mask(S(T_j), A(T_j))$, refines the raw actions generated by the actor network by enforcing action validity and utilizing domain-specific knowledge, thereby improving policy performance. Additionally, policy guidance incorporates the MILP solver discussed earlier to provide optimal actions based on current and future information. These optimal actions are stochastically introduced during RL training into the replay buffer (i.e., tossing a biased coin) to mix high-quality actions given a deterministic trajectory with exploratory actions).

\subsubsection{Action Masking}
Action masking ensures that the policy actions generated by the actor network are feasible during DDPG training. Findings from \cite{huang2020closer,kanervisto2020action} confirm that differentiable action masking does not interfere with the policy gradient backpropagation process. As a result, the learning process remains effective, while the imposed constraints on the action space prevent the policy from exploring invalid actions, thereby improving training efficiency and optimizing resource usage. 
\begin{algorithm}[t]
    \KwIn{$\textit{state: }S(T_j), \textit{action: } A(T_j) $}
    \KwOut{Masked action: $\MaskAction$} 
\small
Initializing: $\PowerNeed \gets [\PowerNeed(C_i, T_j)]_{C_i \in \mathbf{C}}$; \quad
$\ReTime \gets [\ReTime(\phi(C_i, T_j))]_{C_i \in \mathbf{C}};$
$\epsilon \gets 10^{-5}$; \quad
$C^{max} \gets [C^{max}_i]_{C_i \in \mathbf{C}}$; \quad
$C^{min} \gets [C^{min}_i]_{C_i \in \mathbf{C}}$

\tcp{Mask 1: Set action = 0 if no car is connected}
    $ \MaskAction \gets \frac{\ReTime}{\ReTime + \epsilon} \times A(T_j)$\
    
    \tcp{Mask 2: Stop charging when required SoC is reached for uni-directional chargers}
    $ \MaskAction_{tmp} \gets \MaskAction$; \quad
    $\MaskAction[\textit{uniIdx}] \gets \min(\MaskAction_{tmp}, \frac{\PowerNeed}{\delta})[\textit{uniIdx}]$

    \tcp{Mask 3: Enforce charging to the req. SoC before departure. }
    $\overline{\Power(T_j)} \gets \frac{ \PowerNeed- (\ReTime - 1) \times C^{max} \times \delta }{\delta}$\
    $\overline{\Power(T_j)} \gets \min(\overline{\Power(T_j)}, C^{max})$; 
    $\MaskAction \gets \max(\MaskAction, \overline{\Power(T_j)})$\
    
    \tcp{Mask 4:  Bidirectional chargers discharge to req. SoC by departure.}
    $\Power^*(T_j) \gets \frac{\PowerNeed- (\ReTime - 1) \times C^{min} \times \delta }{\delta}$\
    $\Power^*(T_j) \gets \max(\Power^*_t, C^{min})$\
    
    $\MaskAction_{tmp}\gets \MaskAction$; \quad
    $ \MaskAction[\textit{biIdx}] \gets \min(\MaskAction_{tmp}, \Power^*_t)[\textit{biIdx}]$\

    \tcp{Mask 5: Power improvement strategy}
    $ \textit{powerGap} \gets \Building(T_j) - \PrdPeak(T_j)$\
    $ \textit{canIncrease} \gets \textit{ReLU}\left(\min\left(\frac{\PowerNeed}{\delta}, C^{max}\right) - \MaskAction \right)$
    
    $ \textit{toImprove} \gets \min\left(\textit{ReLU}(\textit{powerGap} - \sum \MaskAction), \sum \textit{canIncrease}\right)$
    
    $ \MaskAction \gets \MaskAction + \frac{\textit{toImprove} \times \textit{canIncrease}}{\sum(\textit{canIncrease}) + \epsilon}$

    \tcp{Mask 6: Do not discharge below building load}
    $ \textit{toImprove} \gets \max(-\Building(T_j) - \sum(\MaskAction), 0)$
    $ \textit{negAction} \gets \textit{ReLU}(\MaskAction \times -1) \times -1$
    
    $ \MaskAction \gets \MaskAction +  \frac{\textit{toImprove} \times \textit{negAction}}{\sum(\textit{negAction}) + \epsilon}$

    \caption{Action Masking: $\Mask(S(T_j), A(T_j))$.} 
    \label{alg: action_masking}
\end{algorithm} 

This procedure takes the RL raw action $A(T_j)$, an array of charging power $[P(C_i, T_j)]_{C_i\in\mathcal{C}}$ for all chargers, processes it through the following masking steps, and outputs the masked actions $A'$. Before starting the procedure, we need to obtain the following state features: the remaining power needed to reach the required SoC for all connected EVs ($\PowerNeed$), the time remaining for each EV ($\ReTime$), and the maximum ($C^{\max}$) and minimum ($C^{\min}$) power of all chargers (line 1 in Algorithm~\ref{alg: action_masking}). Also, for our case, since we work with both unidirectional and bidirectional, we denote ${\it uniIdx}$ and ${\it biIdx}$ as the indices for unidirectional and bidirectional chargers, respectively. All of the masking techniques referenced below are from Algorithm~\ref{alg: action_masking}.
\begin{itemize}[leftmargin=*]
    \item \textbf{Mask 1.} 
    We set the charging power $P(C_i, T_j)$ of charger $C_i$ to 0 if no EV is connected, i.e., $\ReTime(\phi(C_i, T_j))=0$. (line 2)
     \item \textbf{Mask 2.} Overcharging unidirectional chargers is not beneficial since excess energy cannot be discharged. Thus, we limit the charging power to ensure the SoC of EVs connected to a unidirectional charger remains within their required SoC. 
    For each connected EV, the actions are masked to the minimum of the current charging power and the power needed to reach its required SoC $\left(\frac{\PowerNeed}{\delta}\right)$ (line 3). 
     
    \item \textbf{Mask 3.} 
    If necessary, we want to adjust actions such that it forces charging to the required SoC before departure to minimize missing SoC, as in Equation~(\ref{eq: soc_penalty}).
    We compute the critical power $\overline{\Power^*(T_j)}$, which is the minimum power required for all chargers at time $T_j$ to reach the required SoC of the connected EVs before departing (assuming maximum power $C^{max}$ is utilized in subsequent time slots). The raw action is adjusted if it falls below this value, especially in time slots leading up to the EV's departure (line 4).    
    \item \textbf{Mask 4.} 
    This mask is symmetrical to Mask 3 for force discharging.
    Overcharging bidirectional EVs is only advantageous if excess energy can be discharged during peak hours, but there is no benefit to overcharging just before departure. Using this mask, we force discharge EVs connected to bidirectional chargers, which have excess energy, and they reach the required SoC by departure.  Here, $\Power^*(T_j)$ denotes the minimum power to discharge for all chargers $C_i \in \mathcal{C}$ at time $T_j$ to guarantee EV can reduce to required SoC when departing (assuming the maximum discharging power $C^{min}$ is utilized subsequently) (lines 5, 6).    
    \item \textbf{Mask 5.} 
    We increase charging power while ensuring the masked action stays within the estimated peak power $\PrdPeak(T_{j})$. This aims to charge EVs as much as possible towards their required SoC without raising demand charges, thereby avoiding forced charging just before departure, which could elevate peak power. 
    We calculate the ``power gap'' between estimated peak power and current building power, $\PrdPeak(T_j) - \Building(T_j)$. If the current power sum ($\Building(T_{j-1}) + \sum_{C_i \in \mathcal{C}} P(C_i, T_{j-1})$) is below this ``power gap'', we boost the current actions using the available ``power'' gap, constrained by $\min \left(\frac{\PowerNeed}{\delta}, C^{max}\right)$. (lines 7 to 9). 
    \item \textbf{Mask 6.} We adjust the discharging power to prevent cumulatively discharging below the current building power $\Building(T_j)$, to satisfy Constraint~\ref{eq:building_power} by reducing the discharging power based on the current actions (lines 10 to 11).
\end{itemize} 
All of the action masking procedures utilize array computations and differentiable operations, such as ReLU \cite{rasamoelina2020review} and maximum/minimum operations, and the PyTorch framework~\cite{paszke2017automatic}. 

\subsubsection{Policy Guidance with MILP Solver}
Note that for a fixed sample, i.e., a fixed set of EV arrivals and departures, the V2B problem can be modeled as a single-shot mathematical program, i.e., a mixed-integer linear program (MILP), which can solved efficiently (at least, for our problem size) to retrieve the optimal actions. The objective of the MILP is maximizing the multi-objective weighted sum of the total rewards (detailed in Equations~\ref{eq: billing}, \eqref{eq: soc_penalty}), and the other properties of the V2B problem can be encoded as constraints. The fixed sample of arrivals and departures can be extracted from historical data. Naturally, this modeling paradigm does not solve the V2B problem in general---EV arrivals and departures are not known ahead of time---however, this strategy provides a set of optimal actions that the learning module can \textit{learn to imitate}. For our use case, the MILP problem can be solved reasonably fast. For example, for a planning horizon of a day with 15 cars, the problem size averages 800 variables and 1400 constraints and takes $0.05$ seconds to solve. 

We integrate a MILP solver based on CPLEX~\cite{cplex2009v12} as a policy guidance subroutine~\cite{pmlr-v28-levine13} in the RL training process. The solver, given the current state and future events, provides optimal charging actions.
{\color{black} Each training dataset contains complete episode data, enabling the MILP solver to account for future dynamics. During RL training, it generates optimal actions based on the current state and full future information of the episode (i.e., a full-month billing period).} The solver is stochastically triggered, and its outputs are added to the replay buffer with a predefined coefficient, $\policyGuidanceRate$ (see Algorithm~\ref{alg:DDPG} in the appendix). The next optimal action is computed as $\mathit{MILP(S(T_j), {\it remainEpisode})}$, considering factors such as EV arrivals, SoC requirements, and building power.
By blending MILP-generated actions with those from the RL actor network, the agent explores a more effective action space, improving its ability to handle large continuous action spaces and long-term rewards.

%
%
%

\subsubsection{Actor-Critic Network Structure} 
Both the actor and critic networks are fully connected, having two hidden layers with 96 neurons each. Both feature a ReLU activation layer at the end. The critic network outputs a single Q-value estimate, while the actor network outputs the action, which represents the charging power of each charger.
To enhance convergence and improve generalization, we normalize all state variables to be within $[0, 1]$ before feeding them into neural networks. Time slot $T_j$ is normalized by division with the number of time slots in a day ($\frac{24}{\delta}$), while power-related variables such as building power $\Building(T_j)$, estimated peak power $\PrdPeak(T_j)$ are scaled by their respective statistical values from training data. Furthermore, we normalize the energy capacity $CAP(V)$ of each car by division with the maximum capacity among EVs, $\max(CAP(V))$.
For the action $A(T_j)=[P(C_i, T_j)]_{C^{i}\in \mathcal{C}}$, we constrain the output within the range $[-1, 1]$ using the $\tanh$ activation function. It is finally translated into the 
charging power range $[C_i^{min}, C_i^{max}]$ by scaling the value using a constant factor.


\subsubsection{Heuristics and Action Post Processing}


To enhance the ease of learning in this complex decision space, we use the RL model on weekdays and the peak hours of TOU price within each billing period (for both training and inference). For off-peak hours and weekends, we use a heuristic based on the least laxity task scheduling algorithm (described in \Cref{sec:experiments_and_results}) to ensure EVs achieve the required SoC before departure, calculating the minimum charge needed for each time slot. Off-peak hours offer lower electricity prices, allowing for higher EV charging rates, and are excluded from demand charge calculations, making heuristics effective for optimization. Similarly, weekends see fewer EV arrivals and lower power demand, with Transmission System Operators excluding them from demand charge assessments. 
Following the EV manufacturer guidelines, we limit charging to SoC boundaries by clipping the actions of the learned policy within $[SoC^{\text{min}}, SoC^{\text{max}}]$ through post-processing to satisfy Constraints~(\ref{eq:soc_min}) and~(\ref{eq:soc_max})

\subsection{Inference}
During execution, our RL-based policy, which is a trained actor network with the action masking procedure, operates at $\delta$ time intervals to determine the charging power for all chargers. At each time slot, the state features are generated from data captured from the environment, including charger status (connected EV's current SoC, expected departure time, and SoC), the building's current power and charging rate limits.
While we use the estimated peak power $\hat{P}^{max}$ as the state feature based on training samples, as shown in~\Cref{fig:pipeline}, it can be replaced by any data-driven forecasting or prediction model. Then, we input all the normalized state features, as described in~\Cref{ssec:MDP}, into the trained RL model to get the charging actions for the next time interval.



\section{Experiments and Analysis}
\label{sec:experiments_and_results}
To demonstrate the performance of our proposed approach, we use data collected from our \nissan{}'s research laboratory. We evaluate our approach against several baselines in terms of total bill and peak shaving (demand charge savings).

\noindent \textbf{Data Collection}
We collected real-world data from \nissan{}'s research laboratory in Santa Clara, California, including building power, EV charger usage, and EV telemetry, over a nine-month period from May 2023 to January 2024. To model the distributions of EV arrivals, SoC requirements, and building power fluctuations, we used Poisson distribution based on historical data. Characteristics of the datasets are shown in~\Cref{ss:appendix_figures}.
{\color{black} The number of EVs arriving at the office on weekdays varies daily, illustrating the inherent uncertainties. Arrival and departure hours relative to SoC are depicted in~\Cref{fig:car_distributions} in the appendix, which also presents the distribution of peak power draw and corresponding hours. Main environment parameters are provided in Table~\ref{tab:simulation_parameters} (appendix).} We sampled 1000 billing episodes for each month. 

\noindent \textbf{Downsampling.} 
We found that increasing training samples beyond a certain limit raised computational demands and worsened performance (see ablation study in~\Cref{ssec:ablation}). To address this, we applied \textit{k}-means clustering \cite{ikotun2023k} with \( k=5 \), using optimal demand charges from the MILP solution to select 60 training samples and 50 testing samples per cluster, ensuring exclusivity.   
{\color{black} As shown in \Cref{tab:training_testing_data} (appendix), the training and testing datasets span nine months, capturing variations in daily EV arrivals, peak building loads. Daily arrivals range from 6.87 (August) to 20.36 (December), reflecting seasonal demand shifts, while monthly peak building loads vary from 116.49 kW (December) to 221.02 kW (August), demonstrating diverse energy consumption patterns affecting charging strategies.}  

\noindent \textbf{Estimated Peak Power.} To enhance training efficacy, we split the monthly dataset into daily episodes for the model to learn from varying weekday conditions. We include a monthly peak power estimate for each month as an input feature derived from optimal action sequences generated by the MILP solver, using the lower bound of the 99\% confidence interval from training data as a conservative demand charge estimate. This input feature is further tuned during RL training. 

\noindent \textbf{Hyperparameter Tuning.} 
Hyperparameter tuning is performed on the parameters outlined in Table \ref{tab:hyperparameters} in the Appendix, which also shows the parameters of the best models selected for each of the nine months. 
To evaluate the model's performance, we employ a 3-fold cross-validation approach, dividing the 60 monthly training samples into 40 samples for training and 20 samples for evaluation. 

\noindent \textbf{Baseline Approaches.}
{\color{black} We transform training data into input samples for our digital twin/simulator, Optimus~\cite{JP2024}, which simulates the EV charging scenario. To evaluate our RL approach, we compare it with an optimal oracle, a real-world charging baseline, and several heuristics. Brief baseline descriptions are provided here, with details in~\Cref{ss:appendix_approach}.}

%
\begin{itemize}[leftmargin=*]
    \item {\bf Optimal MILP Solver (MILP)}: We model deterministic sequences of EV arrivals and departures and solve the problem using the MILP formulation with IBM ILOG CPLEX Optimization Studio~\cite{cplex2009v12}. \textit{The results serve as an upper bound for comparison, as they utilize an oracle for optimality.}
    \item {\bf Fast Charge (FC)}: This approach simulates current real-world charging procedures, charging all connected EVs as quickly as possible to $\SOCMAX$.
    \item {\bf Trickle Charging (Trickle)}: The trickle charging approach utilizes the trickle charging rate, defined as the minimum required charge at each time slot: $P(C_i, T_j) = \PowerNeed(C_i, T_j)/\ReTime(C_i, T_j)$, to charge all EVs until they reach their required SoC. 
\item \textbf{Trickle Least Laxity First (T-LLF)}: We define the Trickle LLF algorithm (detailed in the Appendix) based on the Least Laxity First approach, a dynamic priority-driven method for scheduling multiprocessor real-time tasks~\cite{leung1989new}. 
In EV charging, we define laxity as the difference between the remaining time before departure and the time required to reach the desired SoC at a constant charging rate~\cite{xu2016dynamic}. 
At each time slot, we compute the ``power gap'' (as $\PrdPeak(T_j) - \Building(T_j)$), using the estimated peak power and the current building power.
This power gap is allocated to all EVs by distributing the trickling charger rate 
to those prioritized by their laxity. 
\item {\bf Trickle Early Deadline First (T-EDF)}: We propose the Trickle EDF algorithm in a similar manner to Trickle LLF, with the only difference being the prioritization method. Trickle EDF follows the Early Deadline First approach (based on time of departure of an EV), which was originally designed as a dynamic scheduling algorithm for real-time systems~\cite{stankovic_EDF}. 
    \item {\bf Charge First Least Laxity First (CF-LLF)}: We compute the available ``power gap'', as in Trickle LLF. 
    Then we calculate the sum of the trickle charging rates for all EVs at the current time slot; if this sum is less than the available ``power gap'', we have capacity for overcharging.  We first assign the charging rate for all EVs to be their trickle charging rates, and then, we charge EVs connected to bi-directional chargers to reach their maximum SoC, following the reverse order of their laxity until the power gap is consumed. 
    If the trickle sum exceeds the power gap, bidirectional EVs are discharged, also based on reverse laxity, to fill the negative gap before resuming the trickle charging.  See \Cref{alg:charge_first_llf} in the appendix.

\item {\bf Charge First Deadline First (CF-EDF)}: This follows the same procedure as Charge First LLF but utilizes a different prioritization metric, focusing on the remaining time before EV departure. 



    \color{black}
    \color{black}
\end{itemize}

\subsection{Results}
\label{ssec:results}

\pgfplotstableread[col sep=comma]{total_bill_3.csv}{\totalbilltable} 
\begin{table*}[h]
\centering
\caption{Total Bill on Test Set (Lower is Better). Best Values in Bold. MILP Provides the Optimal Solution with Oracle Input. (Peak Shaving Results is shown in \Cref{table:peak_shaving_demand} in the Appendix.) 
}
\pgfplotstabletypeset[
    precision=1,        
    create on use/MAY2/.style={
        create col/assign/.code={%
            \edef\entry{\thisrow{MAY_mean}$\pm$\thisrow{MAY_std}}
            \pgfkeyslet{/pgfplots/table/create col/next content}{\entry}
        }
    },
    create on use/JUN2/.style={
        create col/assign/.code={%
            \edef\entry{\thisrow{JUN_mean}$\pm$\thisrow{JUN_std}}
            \pgfkeyslet{/pgfplots/table/create col/next content}{\entry}
        }
    },
    create on use/JULY2/.style={
        create col/assign/.code={%
            \edef\entry{\thisrow{JULY_mean}$\pm$\thisrow{JULY_std}}
            \pgfkeyslet{/pgfplots/table/create col/next content}{\entry}
        }
    },
     create on use/AUG2/.style={
        create col/assign/.code={%
            \edef\entry{\thisrow{AUG_mean}$\pm$\thisrow{AUG_std}}
            \pgfkeyslet{/pgfplots/table/create col/next content}{\entry}
        }
    },
    create on use/SEP2/.style={
        create col/assign/.code={%
            \edef\entry{\thisrow{SEP_mean}$\pm$\thisrow{SEP_std}}
            \pgfkeyslet{/pgfplots/table/create col/next content}{\entry}
        }
    },
    create on use/OCT2/.style={
        create col/assign/.code={%
            \edef\entry{\thisrow{OCT_mean}$\pm$\thisrow{OCT_std}}
            \pgfkeyslet{/pgfplots/table/create col/next content}{\entry}
        }
    },
     create on use/NOV2/.style={
        create col/assign/.code={%
            \edef\entry{\thisrow{NOV_mean}$\pm$\thisrow{NOV_std}}
            \pgfkeyslet{/pgfplots/table/create col/next content}{\entry}
        }
    },
     create on use/DEC2/.style={
        create col/assign/.code={%
            \edef\entry{\thisrow{DEC_mean}$\pm$\thisrow{DEC_std}}
            \pgfkeyslet{/pgfplots/table/create col/next content}{\entry}
        }
    },
    create on use/JAN2/.style={
        create col/assign/.code={%
            \edef\entry{\thisrow{JAN_mean}$\pm$\thisrow{JAN_std}}
            \pgfkeyslet{/pgfplots/table/create col/next content}{\entry}
        }
    },
    create on use/TOTAL2/.style={
        create col/assign/.code={%
\edef\entry{\thisrow{TOTAL}$\pm$\thisrow{TOTAL_std}}
            \pgfkeyslet{/pgfplots/table/create col/next content}{\entry}
        }
    },
    column type=c,
    columns/MAY2/.style={column name={MAY}, string type,column type=c}, 
    columns/JUN2/.style={column name={JUN}, string type,column type=c}, 
    columns/JULY2/.style={column name={JULY},string type,column type=c}, 
    columns/AUG2/.style={column name={AUG}, string type,column type=c}, 
    columns/SEP2/.style={column name={SEP}, string type,column type=c}, 
    columns/OCT2/.style={column name={OCT}, string type,column type=c}, 
    columns/NOV2/.style={column name={NOV},string type,column type=c}, 
    columns/DEC2/.style={column name={DEC}, string type,column type=c}, 
    columns/JAN2/.style={column name={JAN}, string type,column type=c},
    columns/TOTAL2/.style={column name={TOTAL}, string type,column type=c},
    fixed,fixed zerofill,     
    columns/Policy/.style={
        string type,
      },
    columns ={Policy, MAY2, JUN2, JULY2, AUG2, SEP2, OCT2, NOV2, DEC2, JAN2},
    every head row/.style={before row=\toprule, after row=\midrule},
    every last row/.style={after row=\bottomrule},
    every first row/.style={before row={\rowcolor[gray]{.8}}},
    every row 1 column 1/.style={postproc cell content/.style={@cell content={\textbf{##1}}}}, 
    every row 2 column 2/.style={postproc cell content/.style={@cell content={\textbf{##1}}}}, 
    every row 1 column 3/.style={postproc cell content/.style={@cell content={\textbf{##1}}}}, 
    every row 1 column 4/.style={postproc cell content/.style={@cell content={\textbf{##1}}}}, 
    every row 1 column 5/.style={postproc cell content/.style={@cell content={\textbf{##1}}}}, 
    every row 1 column 6/.style={postproc cell content/.style={@cell content={\textbf{##1}}}}, 
    every row 1 column 7/.style={postproc cell content/.style={@cell content={\textbf{##1}}}}, 
    every row 1 column 8/.style={postproc cell content/.style={@cell content={\textbf{##1}}}}, 
    every row 1 column 9/.style={postproc cell content/.style={@cell content={\textbf{##1}}}}, 
    every row 1 column 10/.style={postproc cell content/.append style={@cell content={\textbf{##1}}}, zerofill, precision=2,}, 
    ] {\totalbilltable}
\label{table:monthly_total_bill}
\end{table*}

\pgfplotstableread{
RL_mean RL_std RLCluster_mean RLCluster_std RLRandom_mean RLRandom_std RLMoreFeature_mean RLMoreFeature_std RLNoE_mean RLNoE_std RLNoP_mean RLNoP_std RLNoA_mean RLNoA_std Random_mean Random_std PPO_mean PPO_std
20471.9 137 20494.8 174 20511.6 184 20594.1 181 21130.2 214 21157.0 204 21273.7 209 21627.3 180 23385.3 263
}{\ablationtable}

\begin{table*}[ht]
\centering
\caption{Ablation Results for the Total Bill Over Three Months (Lower is Better).}
\pgfplotstabletypeset[
    precision=1, 
    create on use/RL/.style={
        create col/assign/.code={%
            \edef\entry{\thisrow{RL_mean}$\pm$\thisrow{RL_std}}
            \pgfkeyslet{/pgfplots/table/create col/next content}{\entry}
        }
    },
    create on use/RLCluster/.style={
        create col/assign/.code={%
            \edef\entry{\thisrow{RLCluster_mean}$\pm$\thisrow{RLCluster_std}}
            \pgfkeyslet{/pgfplots/table/create col/next content}{\entry}
        }
    },
    create on use/RLRandom/.style={
        create col/assign/.code={%
            \edef\entry{\thisrow{RLRandom_mean}$\pm$\thisrow{RLRandom_std}}
            \pgfkeyslet{/pgfplots/table/create col/next content}{\entry}
        }
    },
    create on use/RLMoreFeature/.style={
        create col/assign/.code={%
            \edef\entry{\thisrow{RLMoreFeature_mean}$\pm$\thisrow{RLMoreFeature_std}}
            \pgfkeyslet{/pgfplots/table/create col/next content}{\entry}
        }
    },
    create on use/RLNoE/.style={
        create col/assign/.code={%
            \edef\entry{\thisrow{RLNoE_mean}$\pm$\thisrow{RLNoE_std}}
            \pgfkeyslet{/pgfplots/table/create col/next content}{\entry}
        }
    },
    create on use/RLNoP/.style={
        create col/assign/.code={%
            \edef\entry{\thisrow{RLNoP_mean}$\pm$\thisrow{RLNoP_std}}
            \pgfkeyslet{/pgfplots/table/create col/next content}{\entry}
        }
    },
    create on use/RLNoA/.style={
        create col/assign/.code={%
            \edef\entry{\thisrow{RLNoA_mean}$\pm$\thisrow{RLNoA_std}}
            \pgfkeyslet{/pgfplots/table/create col/next content}{\entry}
        }
    },
    create on use/Random/.style={
        create col/assign/.code={%
            \edef\entry{\thisrow{Random_mean}$\pm$\thisrow{Random_std}}
            \pgfkeyslet{/pgfplots/table/create col/next content}{\entry}
        }
    },
    create on use/PPO/.style={
        create col/assign/.code={%
            \edef\entry{\thisrow{PPO_mean}$\pm$\thisrow{PPO_std}}
            \pgfkeyslet{/pgfplots/table/create col/next content}{\entry}
        }
    },
    columns/RL/.style={column name={RL (Ours)}, string type, column type=c},
    columns/RLCluster/.style={column name={\rlcluster}, string type, column type=c},
    columns/RLRandom/.style={column name={\rlrandom}, string type, column type=c},
    columns/RLMoreFeature/.style={column name={\rlmorefeature}, string type, column type=c},
    columns/RLNoE/.style={column name={\rlnoe}, string type, column type=c},
    columns/RLNoP/.style={column name={\rlnop}, string type, column type=c},
    columns/RLNoA/.style={column name={\rlnoa}, string type, column type=c},
    columns/Random/.style={column name={\random}, string type, column type=c},
    columns/PPO/.style={column name={{\bf PPO}}, string type, column type=c},
    every head row/.style={before row=\toprule, after row=\midrule},
    every last row/.style={after row=\bottomrule},
    columns={RL, RLCluster, RLRandom, RLMoreFeature, RLNoE, RLNoP, RLNoA, Random}, 
]{\ablationtable}
\label{table:ablation}
\end{table*}

\noindent We evaluate all approaches using two metrics:  1) \textbf{Total Bill:} The sum of electricity cost and demand charge over the billing period, computed by Eq.~(\ref{eq: billing}) and  2) \textbf{Peak Shaving:} It is the difference in demand charge between (i) the building's power usage (without any charging) and (ii) by adding charging the EVs under the respective policies. Positive values indicate that the policy reduced the demand charge by controlling the charging actions. 
Additionally, missing SoC—the energy shortfall between required and actual SoC at departure—is critical in the V2B problem. Our RL model, with action masking, ensures all EVs reach their required SoC before departure by applying force charging and discharging in \textbf{Mask 2} and \textbf{Mask 3}. For fairness, these force procedures are applied across all proposed heuristics, effectively minimizing missing SoC. Therefore, we do not report this metric separately.


We assess the RL model's long-term performance from May 2023 to January 2024, comparing it against baseline approaches on 50 testing samples. 
{\color{black}~\Cref{table:monthly_total_bill} compares the total bill over nine months across different policies. While MILP offers an oracle-based optimal solution, it is impractical for real-world use and serves as a performance upper bound. The results show that the trained RL model consistently achieves the lowest total bills from May 2023 to January 2024 (except June 2023), outperforming other real-time policies in eight of the nine months and significantly reducing costs compared to the real-world Fast Charge procedure as detailed in Table~\ref{table:monthly_total_bill}.
Additionally, heuristic approaches using the First Charge logic, like First Charge LLF or EDF, consistently result in relatively lower total bills and demand charges compared to other heuristics. This indicates that the First Charge approach is effective in balancing the charging and discharging process, offering better overall performance across all heuristics.
\color{black}
\Cref{table:peak_shaving_demand}  in Appendix~\ref{appendix:results} illustrates the peak shaving performance in all approaches, showing that our RL approach achieved peak shaving in six months (indicated by positive values), demonstrating its effectiveness in reducing demand charges by charging EV.
}

\subsection{Ablation Study}
\label{ssec:ablation}

We evaluate the contributions of key techniques in our approach through ablation. For the ablation studies, we trained RL models on monthly samples of three months, May to July 2023, and tested their performance on the total bill. The ablations explored are: 
1) \rlcluster, RL training with more (500) training samples. 
2) \rlrandom, RL training using 60 randomly selected samples from 1000 generated samples.
3) \rlmorefeature, RL models trained using the complete set of 100 state features defined in ~\Cref{ssec:MDP}. 
4) \rlnoe, RL training where the monthly estimated peak power is set to 0, removing the influence of long-term peak power estimation. 
5) \rlnop, RL training without policy guidance. 
6) \rlnoa, RL training without action masking, except for forced charging and discharging (Masks 2 and 3), which are retained to minimize missed SoC. 
7) \random, where actions are randomly selected instead of using a trained actor network, followed by action masking.
We present the sum of the monthly total bills from May to July 2023 for all approaches in the ablation study in Table~\ref{table:ablation} and~\Cref{ss:appendix_approach}. 
 
We evaluate the impact of downsampling using k-means clustering to generate 60 training samples from a pool of 1000. The \rlcluster approach, which uses 500 samples, showed no improvement in performance but increased computational burden during training. We also tested \rlrandom, where samples were randomly selected instead of clustered, resulting in a performance drop. These findings confirm that our downsampling method maintains RL performance while improving efficiency.

We then examine the \rlmorefeature approach, which performs worse, suggesting that condensing state features with domain-specific knowledge improves training and leads to better outcomes. 
The \rlnop\  approach, which removes policy guidance, results in decreased performance, highlighting its importance in optimizing actions during training. This guidance narrows down the action exploration space, directing the model toward better solutions.
 
The \rlnoe\  approach shows worse results, highlighting the importance of accurate long-term peak power estimation during training. This value is used in action masking to improve the charging actions without increasing the monthly peak power and influences the reward function by penalizing actions that raise peak power. When set to $0$, the RL model fails to converge to a good global optimum, emphasizing the critical role of peak power estimation in achieving optimal performance. 

Training without the action masking procedure in \rlnoa \ leads to a significant performance drop, demonstrating its importance in improving RL performance. This also highlights the challenge of training RL models with 15 chargers in a continuous action space. Action masking incorporates heuristics to guide actions, resulting in significant improvements.

To assess the impact of the actor network, we replaced it with a random policy in the \random approach, where random charging actions are generated before applying action masking. Its poor performance highlights that action masking alone is insufficient, emphasizing the actor network’s critical role in achieving optimal outcomes. While all proposed heuristics (except FC and Trickle) adhere to action masking constraints, including forced charging and power allocation based on estimated peak power, the RL approach consistently outperforms them, reinforcing the importance of the actor network.

\section{Conclusion}\label{sec:conclusion}
We propose an RL-based approach to address V2B challenges in smart buildings by optimizing charging power for heterogeneous (mixed-mode) EV chargers. The goal is to minimize overall costs, including energy bills and demand charges, while ensuring EVs reach their required SoC. Our solution addresses key challenges such as multi-agent decision-making, centralized control of up to 15 chargers, and continuous charging power adjustments, all aimed at minimizing the total energy bill over a month.
We evaluate our approach against heuristic algorithms in simulated V2B scenarios with real-world data from an EV manufacturer. Results show that our trained models effectively manage online EV charging, reducing monthly total bills while meeting SoC requirements. 
\section{Acknowledgement}

This material is based upon work sponsored by the National Science Foundation (NSF) under Award Numbers 1952011 and 2238815 and by Nissan Advanced Technology Center-Silicon Valley. Results presented in this paper were obtained using the Chameleon Testbed supported by the NSF. Any opinions, findings, conclusions, or recommendations expressed in this material are those of the authors and do not necessarily reflect the views of the NSF or Nissan. 



\bibliographystyle{ACM-Reference-Format} 
\balance

\bibliography{main}
 
\clearpage
\appendix
\section{Appendix}

\begin{table*}[ht]
\centering
\footnotesize
\caption{Comparison of state-of-the-art approaches for EV charging problem with our approach.}
\begin{tabular}
{|p{2.8cm}|p{2cm}|p{2cm}|p{2cm}|p{1.1cm}|p{1.1cm}|p{1.0cm}|p{0.6cm}|p{1.0cm}|}
\hline
    \textbf{Reference} & \textbf{Approach}  & \textbf{Objective} & \textbf{Action Space} & \textbf{Planning Horizon} & \textbf{Discharge} & \textbf{Mobility} & \textbf{Req. SoC}& \textbf{Demand Charge} \\  \hline
\citeauthor{AORC2013}~\cite{AORC2013} 
& Distributed control algorithm & EV charging Fairness Allocation & Continuous power rate of 2 chargers & Single day &  &  & & \\ \hline 

\citeauthor{5986769}~\cite{5986769} 
& Rule-based control & Minimize energy cost and grid energy losses & Continuous power rate & Permanent &  &   & &\\ \hline 

\citeauthor{9409126}~\cite{9409126} 
& Scheduling algorithm & Minimize demand charge, total load variation, and capacity distribution fairness & Continuous power rate of 80 chargers & One month & & \checkmark & &\checkmark  \\ \hline 

\citeauthor{MMN2019}~\cite{MMN2019} 
& Deep Q-learning, Deep Policy Gradient & Minimize building energy cost & Boolean decision for turn on/off 3 devices & Single day &  &   & &\checkmark  \\ \hline 


\citeauthor{SNDJ2020}~\cite{SNDJ2020}
& RL: off-policy value-iteration & Minimize power consumption and unfinished charging & Boolean decision (charge or not) on 50 charger stations & Single day &  &\checkmark &\checkmark  & \\ \hline

\citeauthor{NNM2024}~\cite{NNM2024}
& Deep RL: PPO & Minimize energy bill and satisfy user QoS  & Continuous power rate of an EV and a HVAC & Single day & \checkmark & \checkmark &\checkmark &  \\ \hline   

\citeauthor{ZJS2022}~\cite{ZJS2022} 
& Federated RL: Soft Actor and Critic & Maximize EV user benefits and electricity prices & Continuous power rate of 3 chargers & One week & \checkmark & \checkmark &\checkmark &  \\ \hline

Our Approach & DDPG with action masking and policy guidance & Minimize demand charge, electricity cost  and missing SoC& Continuous power rate of 15 chargers & One month & \checkmark & \checkmark  &\checkmark &\checkmark \\ \hline
\end{tabular}
\label{tab:comparison}
\end{table*}

\subsection{Related Work}

We provide a more detailed review of prior work here.~\Cref{tab:comparison} describes the key papers and summarizes their gaps.

\noindent\textbf{Uncertainty of vehicles and SoC requirements.} Prior work has taken \citeauthor{MJG2015} defines this taking into consideration different mobility aspects such as the arrival/departure time of an EV at/from a charging station, trip history of EVs, and unplanned departure of EVs~\cite{MJG2015}. Empirical studies, such as those by \citeauthor{richardson2011electric}, have analyzed EV charging strategies and their impact on grid stability, which are closely related to V2B systems~\cite{richardson2011electric}. 
The challenges of optimizing V2B systems, especially given the dynamic nature of energy pricing and vehicle usage patterns, have also been addressed by \citeauthor{8274175}, who proposed a demand response framework for smart grids~\cite{8274175}. Additionally, \citeauthor{oconnell2010integration} applied optimization algorithms, such as Mixed Integer Linear Programming (MILP), to integrate renewable energy sources into grid systems~\cite{oconnell2010integration}. 
{\color{black} Other approaches, including meta-heuristics and Model Predictive Control (MPC), have been explored to optimize the smart EV charging process for electric vehicles (EVs), focusing on energy cost and user fairness in single-site or vehicle-to-grid (V2G) systems~\cite{AORC2013, 5986769, 9409126, MJG2015}.} 
However, many of these methods focus on unidirectional chargers and fail to fully account for uncertainty including, vehicle arrivals and departures~\cite{MJG2015}. 

\noindent\textbf{Time of use pricing, demand charge, and long-term rewards.} 
Optimizing V2B is a complex problem, made more complex when the lengths of billing periods set by TSOs are considered. Several approaches~\cite{AORC2013, MMN2019, SNDJ2020} only optimize and plan for single-day horizons. ~\citeauthor{9409126} are able to achieve one-month planning horizons while considering demand charge~\cite{9409126}. However, they assume a homogeneous set of chargers. Thus preventing them from fully realizing the effect of EVs on potential savings.

\noindent\textbf{Heterogeneous chargers and continuous action spaces.} Approaches that solve EV charging without considering the ability of EVs to discharge ignore even more potential savings. However, addressing this introduces further complexity to the system.
\citeauthor{NNM2024} works around the limitations of charger homogeneity by using Deep RL~\cite{NNM2024}. They consider SoC requirements and address the mobility-aware needs of EVs. However, their approach does not consider long-term rewards, instead limiting their planning to a single day. 
Improving upon these initial approaches, \citeauthor{ZJS2022} investigated federated RL for EV charger control, aiming to maximize user benefits~\cite{ZJS2022}, and minimize electricity prices. Their approach explores the continuous action space of charging power and extends their planning horizon to an entire week. While their approach includes both discharging and charging actions, they fail to capture the idea of demand charge into their problem, which is critical in the industrial context. 

\noindent\textbf{Tracking real-world state and transition.}
Existing approaches validate their approaches using simulations that have limited interface with the real world. \citeauthor{9409126} utilizes an existing Adaptive Charging Network (ACN) EV charging testbed along with a mobile application to capture EV telemetry and charger behavior~\cite{9409126}. Thus, they capture the complexity of real charging systems, including battery charging behaviors.

\pgfplotstableread[col sep=comma]{results/car_arrival_per_dow.csv}{\cararrivalsdow}
\pgfplotstableread[col sep=comma]{results/building_dow_box.csv}{\buildingmaxdow}

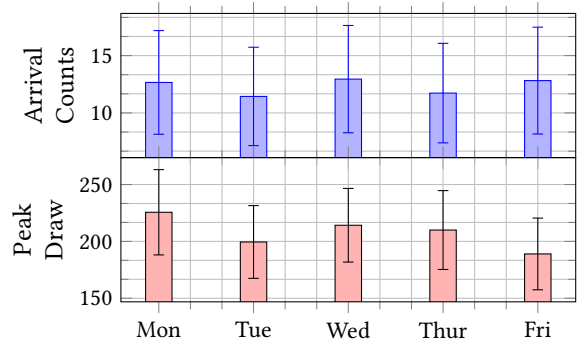
\begin{figure}[H]
\centering
\begin{tikzpicture}
\begin{groupplot}[
    width=0.9\linewidth,
    height=3.5cm,
    group style={
      group size=1 by 2,
      vertical sep=0cm,
    },
]

  \nextgroupplot[
    ybar,
       xticklabels = {},
    legend pos=outer north east,
    ylabel={Arrival\\Counts},
    ylabel style={font=\large, align=center},
    xtick=data,
    xticklabel style={
        rotate=0,
    },
        grid=both,
        ymajorgrids=true,
        yminorgrids=true,
        minor x tick num=2,
        minor y tick num=2,
  ]
\addplot[blue,fill=blue!30!white,error bars/.cd,y dir=both,y explicit,] table[
  x=index, 
  y=mean, 
  col sep=comma,
  y error=std 
] {\cararrivalsdow};

  \nextgroupplot[
    ybar,
       xticklabels = {
       Mon,
       Tue,
       Wed,
       Thur,
       Fri
       },
    legend pos=outer north east,
    ylabel={Peak\\Draw},
    ylabel style={font=\large, align=center},
    xtick=data,
    xticklabel style={
        rotate=0,
    },
        grid=both,
        ymajorgrids=true,
        yminorgrids=true,
        minor x tick num=2,
        minor y tick num=2,
  ]
\addplot[black,fill=red!30!white,error bars/.cd,y dir=both,y explicit,] table[
  x=index, 
  y=mean, 
  col sep=comma,
  y error=std 
] {\buildingmaxdow};

\end{groupplot}
\end{tikzpicture}

\caption{(Top) Arrival counts per day of the week across 8 months. Most of the cars arrive during Wednesday and Friday. (Bottom) Peak building draw per day of the week across 8 months. Mondays and Fridays typically exhibit the highest and lowest power draws.}
\label{fig:arrival_counts}
\Description[<short description>]{<long description>}
\end{figure}

\pgfplotstableread[col sep=comma]{results/training_data_arrival_distribution.csv}{\arrdata}
\pgfplotstableread[col sep=comma]{results/training_data_departure_distribution.csv}{\depdata}
\pgfplotstableread[col sep=comma]{results/weekday_building_hour_peaks_distribution.csv}{\bldgdata}
\pgfplotstableread[col sep=comma]{results/grid_prices.csv}{\griddata}

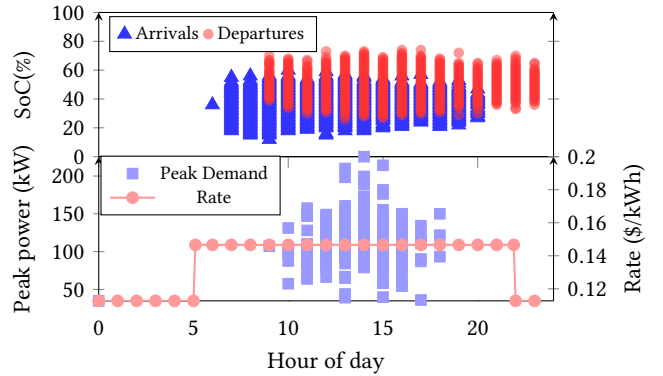
\begin{figure}[H]
\begin{tikzpicture}

\begin{groupplot}[
            group style={
        group size=1 by 2,
        xlabels at=edge bottom,
        xticklabels at=edge bottom,
        vertical sep=0pt,
        },
        width=0.9\linewidth,
        height=3.5cm,
        xlabel=Hour of day,
        xmin=0, xmax=24,
        xtick align=outside,xtick pos=left,
        ytick align=outside,
        ytick pos=left,
        axis y line=left,
        legend style={nodes={scale=0.8, transform shape}, 
                      legend columns=2,
                      at={(0.48, 0.85)},
                      anchor=east,
        },
        ]

\nextgroupplot[ymin=0, ymax=100, ylabel=SoC(\%), ytick align=outside,ytick pos=left,axis y line=left]

   \addplot[only marks,mark=triangle*,blue!80,
   mark size=3pt,
   ] table[x=ahour,y=soc] {\arrdata};
   \addlegendentry{Arrivals}

   \addplot[only marks,mark=*, red!80,
   fill opacity=0.5, draw opacity=0,
   mark size=2pt,
   ] table[x=dhour,y=required_soc_at_depart] {\depdata};
   \addlegendentry{Departures}
   
\nextgroupplot[ylabel={Peak power (kW)}, yticklabel={},
        xtick style={
            /pgfplots/major tick length=0pt,
        },
        legend style={nodes={scale=1.0, transform shape}, 
                      legend columns=2,
                      at={(0.35, 0.85)},
                      anchor=east,
        },
]  
   \addplot[only marks,mark=square*,blue!40] table[x=hour,y=power] {\bldgdata};
   \label{plot:peakdemand}

\end{groupplot}

\begin{groupplot}[
        group style={
             group name=plots,
             group size=1 by 2,
             xlabels at=edge bottom,
             xticklabels at=edge bottom,
             vertical sep=0pt,
        },
        width=0.9\linewidth,
        height=3.5cm,
        xmin=0, xmax=24,
        ytick align=outside,
        ytick pos=right,
        axis y line=right,
        axis x line=none,
        legend style={nodes={scale=0.8, transform shape}, 
                      legend columns=1,
                      at={(0.4, 0.8)},
                      anchor=east,
        },
        ]

\nextgroupplot[ymin=0,ymax=500,
               ylabel={},
               yticklabels={},
]

\nextgroupplot[
ymin=0.11271,ymax=0.2,
               ylabel={Rate (\$/kWh)},
               ]  
\addlegendimage{/pgfplots/refstyle={plot:peakdemand}}\addlegendentry{Peak Demand}
\addplot[thick, mark=*, mark options={solid, color=red!40}, color=red!40] table[col sep=comma, x=hour, y=price] {\griddata};
   \addlegendentry{Rate}

\end{groupplot}
\end{tikzpicture}

\caption{(Top) Distribution of EV hours and SoC for both arrivals and departures across 8 months. (Bottom) Distribution of peak building power draw against the hours of day it was drawn and the TOU rates across the day. The majority of the car arrivals and departure occur within high peak pricing times and all of the peak power draw happens during peak pricing time.}
\Description[<short description>]{<long description>}
\label{fig:car_distributions}
\end{figure}

\subsection{Complementary Figures}
\label{ss:appendix_figures}
The characteristics of the training set are shown in~\Cref{fig:arrival_counts} and~\Cref{fig:car_distributions}.~\Cref{fig:arrival_counts} show the variations in arrival counts (top) and peak building draw (bottom) across the different days of the week across 8 months. Most cars arrive on Wednesdays and Fridays while the peak building draw is the least on Fridays.~\Cref{fig:car_distributions} (top) show distributions of car arrivals and departures hours against the EVs arrival SoC and required SoC upon departure.~\Cref{fig:car_distributions} (bottom) show the distribution of peak power draw and the hour of day. The line in red signifies the TOU rates across the day. All of the cars arrive within peak hours, while the majority of them leave within the peak hours. Finally, all off-the-peak power draws occur at peak hours. The intersection of these arrivals, departures, and peak power draw hours represents the potential space in which our policy can operate.

\textbf{Environment Simulator.} We process these training data into input samples for our digital twin/simulator: Optimums~\cite{JP2024}. We model a digital twin for the target environment and provide several interfaces that allow both simulated and real-world components to leverage our proposed approach. This allows us to investigate how any action or decision can potentially impact the real world. 
There are two main decisions that must be taken when solving the V2B charging problem. (1) charger assignments and (2) charger actions.

\pgfplotstableread[col sep=comma]{charger_assignments.csv}{\chargerassignments}

\begin{table}[!ht]
\centering
\caption{Charger Assignment and tiebreaker comparisons with an MILP policy. Assigning to Bidirectional chargers first and then breaking ties by assigining them to the EV that departs later results in the lowest bill. Lower is better.}
\pgfplotstabletypeset[
    create on use/bill/.style={
        create col/assign/.code={%
            \edef\entry{\thisrow{c}$\pm$\thisrow{d}}
            \pgfkeyslet{/pgfplots/table/create col/next content}{\entry}
        }
    },
    columns ={a, b, bill},
    column type=c,
    columns/a/.style={string type},
    columns/b/.style={string type,column type=c},
    columns/bill/.style={string type,column type=c},
    fixed,
    fixed zerofill,
    precision=2,
    every head row/.style={
        output empty row,
        before row={
            \toprule
            {Assignment} & {Tie Breaker} & {Average Monthly Bill (\$)}\\
        }, 
        after row=\midrule},
    every last row/.style={after row=\bottomrule},
    every row 0 column 0/.style={postproc cell content/.style={@cell content={\textbf{##1}}}}, 
    every row 0 column 1/.style={postproc cell content/.style={@cell content={\textbf{##1}}}}, 
    ] {\chargerassignments}
\label{table:charger_assignment_policies}
\end{table}

\textbf{Charger assignment.} We consider a first-in, first-out policy that assigns EVs to bidirectional chargers first, breaking ties assigning to later departing cars; a comparison of different policy combinations is shown in~\Cref{table:charger_assignment_policies}. We observe that bidirectional charging assignments outperform all other policies. Tie-breaking strategies that prioritize later-departing vehicles show a marginal advantage. While these assignment policies could be further optimized, we chose to follow this heuristic and focus on the second decision problem: determining charger actions.

\textbf{Charger actions.} We provide several policies with our simulator to compare our proposed approach. Charger action policies receive a state of the environment for a particular time and generate actions based on this.

\begin{algorithm}[t]
\setcounter{AlgoLine}{0} %
\small
    \SetAlgoLined
    \KwIn{Initial policy parameters for actor network $\zeta_a$, critic parameters $\zeta_c$, target network parameters $\zeta_a', \zeta_c'$\\
Training parameters: $\mathit{actionNoise}$, $\policyGuidanceRate$, $\mathit{bufferSize}$, $\mathit{batchSize}$, maximum iterations: $M$,  training steps: ${\it trainStep}$; target network update steps: ${\it updateStep}$
}
    \KwOut{Trained policy $\pi_{\zeta_a}$}
    Initialize replay buffer $\Buffer$;  ${\it step}\gets 0$\\
    \For{$1$ \KwTo $M$}{
        Input a sample into simulator to generate initial state $s_0$ 
        
        \For{each time slot $T_j \in \mathcal{T}$}{
        \tcp{Introducing policy guidance stochastically.}
            {\color{black} randomValue $\leftarrow randomBetween(0,1) $
            
            \If
            { randomValue $\leq \policyGuidanceRate$}{
        Get action $A(T_j)$ by rerunning the MILP solver: $A(T_j)\leftarrow\MILP(S(T_j), {\it remainEpisode})$
                }
            \Else{
                Get masked action using current policy,  $\mathit{actionNoise}$: \\
                $A(T_j) \gets \Mask\left(S(T_j), \pi(S(T_j) | \zeta_a)) + \mathit{actionNoise} \right)$
            }}
             State transition $S(T_{j+1})\leftarrow {\it Trans}(S(T_j), A(T_j))$ 

            Get the action reward $R(T_j)\leftarrow {\it Reward}(S(T_j), A(T_j))$
            
            Store transition $(S(T_j), A(T_j), R(T_j), S(T_{j+1}))$ in $\Buffer$
            
            \If{${\it step} \bmod {\it trainStep} == 0$}{
            Sample batch $(S(T_i), A(T_i), R(T_i), S(T_{i+1})$ from $\Buffer$ 

           {\color{black} Get masked actions using target actor network:\\
            $A(T_{i+1})\leftarrow \Mask(S(T_{i+1}), \pi'(S(T_{i+1}) | \zeta_a'))$}  

            Set target $y_i\leftarrow R(T_j) + \gamma Q'(S(T_{i+1}), A(T_{i+1})
            | \zeta_c')$ 
            
            Update critic network by minimizing the loss: $L 
            \leftarrow \frac{1}{N} \sum_i (y_i - Q(S(T_i), A(T_i) | \zeta_c))^2$ 
            
             {\color{black} Get masked actions $A(T_i)$ at $S(T_i)$ using actor network: $A(T_i)\leftarrow \Mask( S(T_i), \pi(S(T_i) | \zeta_a))$ }
           
            Update the actor policy using policy gradient:\\
            $\nabla_{\zeta_a} J \leftarrow \frac{1}{N} \sum_i \nabla_a Q(S(T_i), A(T_i) | \zeta_c) | \nabla_{\zeta_a} \pi(s | \zeta_a) |_{S(T_i)}$}
            
        \If{${\it step} \bmod {\it updateStep} == 0$}{
             Update the target networks: 
            $\zeta_a' \leftarrow \tau \zeta_a + (1 - \tau) \zeta_a'$;\quad 
            $\zeta_c' \leftarrow \tau \zeta_c + (1 - \tau) \zeta_c'$ \\
            }${\it step}\gets {\it step}+1$
        }
    }
    \caption{Improved DDPG with Action Masking and Policy Guidance.}
\label{alg:DDPG} 
\end{algorithm}


\subsection{Additional Details on the Approach}
\label{ss:appendix_approach}

Our proposed approach, as outlined in Algorithm~\ref{alg:DDPG}, is based on the DDPG algorithm~\cite{lillicrap2015continuous}, 
which utilizes an actor network to generate actions. Tuples of state, action, reward, and next state are stored as transitions in the replay buffer (lines 9 to 12). 
During training, we interact with the environment simulator. We provide input from the training dataset.~\Cref{tab:training_testing_data} shows the characteristics of this dataset. The environment simulator abstracts state features for the RL models and manages state transitions based on the function described in Section~\ref{ssec:MDP}. 
In each training iteration, we batch state transitions from the replay buffer for model training (line 13). Specifically, DDPG maintains target networks for both the actor and critic, which are used to generate the next state and compute Q-values essential for calculating the critic loss during training. The critic network is trained using gradient descent by minimizing the mean squared error between predicted Q-values and target Q-values derived from the Bellman equation (lines 14-16). The critic learns Q-values for state-action pairs, which are then used to train the actor network through a policy gradient approach (lines 17 and 18). The updates to the target networks are delayed to stabilize the training process (lines 18-19). 

To improve RL performance in handling the limitations associated with large continuous action spaces and long-term reward optimization, we introduce action masking and policy guidance techniques. 
Action masking, denoted as $\Mask(S(T_j), A(T_j))$, refines the raw actions generated by the actor network by enforcing action validity and utilizing domain-specific knowledge, thereby improving policy performance (lines 9, 14, 17). Meanwhile, policy guidance incorporates the MILP solver to provide optimal actions through $\MILP(S(T_j), {\it remainEpisode})$, based on current and future information (lines 5-9). These optimal actions are stochastically introduced during RL training into the replay buffer, mixing high-quality actions with the raw RL actions to enhance the training transition quality and guide the RL training in a beneficial direction.

\begin{table}[hpt]
\centering
\small
\caption{Hyperparameters and selected values.}
\begin{tabular}{|>{\raggedright}p{1.7cm}|p{4cm}|p{1.3cm}|}
\hline
Parameter & Description & Range \\ 
\hline
Actor network & Number of units at each layer & [96, 96] \\\hline
Critic network & Number of units at each layer & [96, 96] \\\hline
$\Gamma$ & Discount factor for future reward & 1 \\\hline 
Actor\&Critic learning rate &Learning rate for updating actor and critic networks & $10^{-5}$, $10^{-3}$ \\\hline 
{\it bufferSize} & Batch size for fetching transitions from replay buffer & 64 \\\hline 
{\it batchSize} & Size of the replay buffer & $10^6$ \\\hline 
{\it actionNoise} & Noise added during action exploration & normal(0,0.2)\\\hline 
$\policyGuidanceRate$ & Probability to introduce policy guidance & 0.5 or 0.7 \\\hline  
$\lambda_{S}$, $\lambda_{B}$, $\lambda_{D}$ & Penalty coefficients for SoC requirement, bill cost, and demand charge &1,1, 3\\\hline 
Random seed & Random seed for actor and critic network initialization  &0-5\\\hline  
Adjustment of $\PrdPeak$& Lower bound of the 99\% confidence interval for the optimal monthly peak power based on training data & Increased by 0\%, 5\%, 10\% \\\hline  
${\it trainStep}$, ${\it updateStep}$& Training steps and steps per update of target networks& 5, 5\\\hline

\end{tabular}
\label{tab:hyperparameters}
\end{table} 

\begin{table*}[htp]
\centering
\small 
\caption{Training and testing data information for each month.} 
\begin{tabular}{|p{1.0cm}|p{1.5cm}|p{1.7cm}|p{1.7cm}|p{1.0cm}|p{1.0cm}|p{1.5cm}|p{1.7cm}|p{1.7cm}|}
\hline
\multirow{2}{*}{\textbf{Month}} & \multicolumn{5}{p{3.75cm}|}{\textbf{Training (60 Samples)}} & \multicolumn{3}{p{3.75cm}|}{\textbf{Testing (50 Samples)}} \\ 
\cline{2-9} & {Car Arrival Number (per day)} & {Monthly Peak Building Load (kW)} & {Daily Peak Building load (kW)} & {Estimated peak power (kW)} & {Number of Weekdays} & {Car Arrival Number (per day)} & {Monthly Peak Building Load (kW)} & {Daily Peak Building Load (kW)} \\ \hline 
\textbf{MAY} & $9.35\pm2.19$ & $125.89\pm1.72$ & $96.22\pm13.76$ & 119 & 22 & $9.31\pm2.2$ & $125.62\pm1.65$ &$96.18\pm13.73$ \\ \hline

\textbf{JUN} & $11.4\pm2.53$ & $141.04\pm1.83$ & $109.86\pm14.0$ & 125 & 21 & $11.49\pm2.51$ & $140.34\pm2.21$ &$109.75\pm13.97$\\ \hline

\textbf{JUL} & $10.86\pm2.45$ & $148.08\pm1.57$ & $111.79\pm35.39$ & 145 & 20 & $10.97\pm2.4$ & $148.12\pm1.72$ &$111.94\pm35.49$\\ \hline

\textbf{AUG} & $6.87\pm1.97$ & $221.02\pm3.86$ & $160.5\pm27.92$ & 202 & 23 & $6.92\pm1.93$ & $221.17\pm3.77$ &$160.69\pm28.14$\\ \hline

\textbf{SEP} & $14.99\pm2.9$ & $145.91\pm1.48$ & $127.2\pm11.86$ & 143 & 20 & $14.94\pm2.88$ & $146.23\pm1.08$ &$127.2\pm11.85$\\ \hline

\textbf{OCT} & $14.55\pm2.91$ & $186.54\pm1.72$ & $118.58\pm30.12$& 174 & 21 & $14.54\pm2.86$ &$185.17\pm2.77$ &$118.5\pm30.13$ \\ \hline 

\textbf{NOV} & $9.59\pm2.31$ & $144.9\pm3.28$ & $113.09\pm19.18$ & 130 & 19 & $9.57\pm2.29$ & $144.2\pm2.92$&  $113.09\pm19.2$\\ \hline 

\textbf{DEC} & $20.36\pm3.4$ & $116.49\pm2.25$ & $98.72\pm8.0$ & 104 & 16 & $20.32\pm3.35$ & $116.33\pm2.01$ &$98.69\pm7.95$\\ \hline 

\textbf{JAN} & $13.62\pm2.73$ & $137.77\pm2.47$ & $91.56\pm19.46$ & 127 & 21 & $13.53\pm2.76$ & $137.64\pm2.5$ &$91.72\pm19.47$\\ \hline 
\end{tabular}
\label{tab:training_testing_data}
\end{table*}   

\begin{table}[t]
    \centering
    \small
    \caption{Simulation Parameters.}
     \label{tab:simulation_parameters}
    \begin{tabular}{|p{1.3cm}|p{6.3cm}|}
    \hline
    \textbf{Parameter} & \textbf{Value} \\
    \hline
    $\mathcal{C}$ & 15 chargers (5 bi-directional, 10 uni-directional) \\\hline
    ${\it C}^i_{min}, {\it C}^i_{max}$ & 
    [-20 kW, 20 kW] for bi-directional, [0, 20 kW] for uni-directional chargers \\\hline
    $\delta$ & Time interval: 0.25 hours \\\hline
    $\theta_{D}, \theta_{E}(T_j)$ & 9.62 \$/kW (Demand), 0.11271 \$/kWh (off-peak), 0.1466 \$/kWh (peak: 6 a.m.-10 p.m.) \\\hline
    $CAP(V)$ & EV battery capacity: 40 or 62 kWh \\\hline
    $\SOCMIN(V)$, $\SOCMAX(V)$ & Minimum and maximum SoC: 0\% and 90\% of capacity \\\hline
    \end{tabular}
\end{table}

\subsection{Complementary Experimental Results}
\label{appendix:results}
We evaluate our approach on an environment with the parameters shown in~\Cref{tab:simulation_parameters}. We use the same parameters across hyperparameter tuning, training, evaluation, and comparison.

\noindent \textbf{Additional Details on Hyper-Parameter Tuning:} This methodology enables us to monitor and assess the RL model's performance throughout the training process. 
An early stopping procedure is implemented, terminating training if the total reward on the 20 evaluation samples does not improve after a specified number of iterations. Finally, we select the optimal combination of hyperparameters based on the results from the 3-fold cross-validation. The new RL model is then trained using the full set of training samples with the identified best parameter combination. Finally, we test the trained 9 RL models on 50 unseen monthly samples for each month from May 2023 to January 2024 to evaluate their generalization performance. ~\Cref{tab:hyperparameters} show selected values for each hyperparameter after tuning.

\begin{algorithm}[htp]
\caption{Trickle Charging with Least Laxity First.}
\label{alg:llf}
\small
\KwIn{Set of EVs $V$, chargers $\mathcal{C}$, time slots $t$}
\KwOut{Charging schedule for each EV}

\For{each time slot $T_j \in\mathcal{T}$}{
    \For{each EV $\phi(C_i, T_j)$ connected to chargers $\mathcal{C}$}{
        Compute laxity: 
        $L(\phi(C_i, T_j)) = (\DepartureTime(\phi(C_i, T_j)) - t) - \PowerNeed(C_i, T_j) /C_{max}^i$.
    }
    Compute power gap: 
    $\hat{P} = \PrdPeak(T_j) - \Building(T_j)$\;
    Sort EVs by laxity: $L(\phi(C_i, T_j))$ in ascending order\;
    Initialize $[P(C_i, T_j)]_{C^i\in\mathcal{C}}$ as 0\;
        \For{each EV $\phi(C_i, T_j)$ sorted by laxity}{
            \If{$\hat{P} > 0$}{
                Set trickle charging rate: 
                $P(C_i, T_j) = \min({\PowerNeed(C_i, T_j)}/{\ReTime(C_i, T_j)},\hat{P})$ \;

            $\hat{P}\gets \hat{P}-P(C_i, T_j)$\;
            }
        }
    }
\end{algorithm}  

\begin{algorithm}[htbp]
\caption{Charge First with Least Laxity First.}
\label{alg:charge_first_llf}
\small
\KwIn{Set of EVs $V$, chargers $\mathcal{C}$, time slots $t$}
\KwOut{Charging schedule for each EV}

\For{each time slot $T_j \in \mathcal{T}$}{
    \For{each EV $\phi(C_i, T_j)$ connected to $\mathcal{C}$}{
        Compute laxity: 
        $L(\phi(C_i, T_j)) = (\DepartureTime(\phi(C_i, T_j)) - T_j) - {\PowerNeed(C_i, T_j)}/{C_{max}^i}$.
    }
    Compute power gap: 
    $\hat{P} = \PrdPeak(T_j) - \Building(T_j)$\;
    
    Compute sum of trickle rates: 
    $S = \sum_{i} {\PowerNeed(C_i, T_j)}/{\ReTime(C_i, T_j)}$\;

  \If(\tcp*[h]{Overcharging EVs for future discharging}){$S < \hat{P}$}{
        \For{each EV $\phi(C_i, T_j)$ connected to $\mathcal{C}$}{
            Set trickle rate: 
            $P(C_i, T_j) \gets {\PowerNeed(C_i, T_j)}/{\ReTime(C_i, T_j)}$\;
        }
        $\hat{P} \gets \hat{P} - S$\;

        \For{each EV $V=\phi(C_i, T_j)$ connected to bi-directional chargers in reverse laxity order}{
            Charge to maximum SoC:
            $P(C_i, T_j) \gets \min\left(C^{max}_i, \frac{(\SOCMAX(V) - \SOC(V, T_j)) \times CAP(V)}{\Delta t}, \hat{P}\right)$\;

            $\hat{P} \gets \hat{P} - P(C_i, T_j)$\;
        }
    }
    \Else(\tcp*[h]{Discharging EVs to increase power gap}){ 
        \While{$\hat{P} \geq S$ \textbf{and} not all EVs connected to bi-directional chargers are considered}{
            \For{each $V=\phi(C_i, T_j)$ connected to bi-directional chargers in reverse laxity order}{
                \If{$SOC(V,T_j) > \SOCR(V)$}{
                    $P(C_i, T_j) \gets \max\left(C^{min}_i, \frac{(\SOCR(V) - \SOC(V, T_j)) \times CAP(V)}{\delta }\right)$\;

                    $\hat{P} \gets \hat{P} - P(C_i, T_j)$\;
                }
            }
        }
        \For{each $V=\phi(C_i, T_j)$ connected to chargers sorted by reverse laxity}{
            Set trickle rate:
            $P(C_i, T_j) \gets \min\left({\PowerNeed(C_i, T_j)}/{\ReTime(C_i, T_j)}, \hat{P}\right)$\;

            $\hat{P} \gets \hat{P} - P(C_i, T_j)$\;
        }
    }
}
\end{algorithm}

\pgfplotstableread[col sep=comma]{add_ablation_study.csv}{\ablationtable}

\begin{table}[t]
\centering
\caption{Month-wise ablation results for total bill. Lower is better.}
\pgfplotstabletypeset[
    every row 0 column 1/.style={postproc cell content/.style={@cell content={\textbf{##1}}}}, 
    every row 0 column 2/.style={postproc cell content/.style={@cell content={\textbf{##1}}}}, 
    every row 0 column 3/.style={postproc cell content/.style={@cell content={\textbf{##1}}}}, 
    every row 0 column 4/.style={postproc cell content/.style={@cell content={\textbf{##1}}}}, 
    columns ={Policy,MAY,JUN,JULY,Total},
    column type=c,
    columns/Policy/.style={string type},
    columns/MAY/.style={string type,column type=c},
    columns/JUN/.style={string type,column type=c},
    columns/JULY/.style={string type,column type=c},
    fixed,
    fixed zerofill,
    precision=2,
    fixed,fixed zerofill,
    every head row/.style={before row=\toprule, after row=\midrule},
    every last row/.style={after row=\bottomrule},
    ] {\ablationtable}
\label{table:ablation_study}
\end{table}

\pgfplotstableread[col sep=comma]{demand_charge_2.csv}{\totalbilltable} 
\begin{table*}[tbh]
\centering
\caption{Peak shaving on test set, best values in bold. MILP shows the optimal solution given an oracle input. Higher is better.}
\pgfplotstabletypeset[
    precision=1,        
    create on use/MAY2/.style={
        create col/assign/.code={%
            \edef\entry{\thisrow{MAY_mean}$\pm$\thisrow{MAY_std}}
            \pgfkeyslet{/pgfplots/table/create col/next content}{\entry}
        }
    },
    create on use/JUN2/.style={
        create col/assign/.code={%
            \edef\entry{\thisrow{JUN_mean}$\pm$\thisrow{JUN_std}}
            \pgfkeyslet{/pgfplots/table/create col/next content}{\entry}
        }
    },
    create on use/JULY2/.style={
        create col/assign/.code={%
            \edef\entry{\thisrow{JULY_mean}$\pm$\thisrow{JULY_std}}
            \pgfkeyslet{/pgfplots/table/create col/next content}{\entry}
        }
    },
     create on use/AUG2/.style={
        create col/assign/.code={%
            \edef\entry{\thisrow{AUG_mean}$\pm$\thisrow{AUG_std}}
            \pgfkeyslet{/pgfplots/table/create col/next content}{\entry}
        }
    },
    create on use/SEP2/.style={
        create col/assign/.code={%
            \edef\entry{\thisrow{SEP_mean}$\pm$\thisrow{SEP_std}}
            \pgfkeyslet{/pgfplots/table/create col/next content}{\entry}
        }
    },
    create on use/OCT2/.style={
        create col/assign/.code={%
            \edef\entry{\thisrow{OCT_mean}$\pm$\thisrow{OCT_std}}
            \pgfkeyslet{/pgfplots/table/create col/next content}{\entry}
        }
    },
     create on use/NOV2/.style={
        create col/assign/.code={%
            \edef\entry{\thisrow{NOV_mean}$\pm$\thisrow{NOV_std}}
            \pgfkeyslet{/pgfplots/table/create col/next content}{\entry}
        }
    },
     create on use/DEC2/.style={
        create col/assign/.code={%
            \edef\entry{\thisrow{DEC_mean}$\pm$\thisrow{DEC_std}}
            \pgfkeyslet{/pgfplots/table/create col/next content}{\entry}
        }
    },
    create on use/JAN2/.style={
        create col/assign/.code={%
            \edef\entry{\thisrow{JAN_mean}$\pm$\thisrow{JAN_std}}
            \pgfkeyslet{/pgfplots/table/create col/next content}{\entry}
        }
    },
     create on use/TOTAL2/.style={
        create col/assign/.code={%
            \edef\entry{\thisrow{TOTAL}$\pm$\thisrow{TOTAL_std}}
            \pgfkeyslet{/pgfplots/table/create col/next content}{\entry}
        }
    },
    column type=c,
    columns/MAY2/.style={column name={MAY}, string type,column type=c}, 
    columns/JUN2/.style={column name={JUN}, string type,column type=c}, 
    columns/JULY2/.style={column name={JULY},string type,column type=c}, 
    columns/AUG2/.style={column name={AUG}, string type,column type=c}, 
    columns/SEP2/.style={column name={SEP}, string type,column type=c}, 
    columns/OCT2/.style={column name={OCT}, string type,column type=c}, 
    columns/NOV2/.style={column name={NOV},string type,column type=c}, 
    columns/DEC2/.style={column name={DEC}, string type,column type=c}, 
    columns/JAN2/.style={column name={JAN}, string type,column type=c},
    columns/TOTAL2/.style={column name={TOTAL}, string type,column type=c},
    fixed,fixed zerofill,     
    columns/Policy/.style={
        string type,
      },
    columns ={Policy, MAY2, JUN2, JULY2, AUG2, SEP2, OCT2, NOV2, DEC2, JAN2},
    every head row/.style={before row=\toprule, after row=\midrule},
    every last row/.style={after row=\bottomrule},
    every first row/.style={before row={\rowcolor[gray]{.8}}},
    every row 1 column 1/.style={postproc cell content/.style={@cell content={\textbf{##1}}}}, 
    every row 2 column 2/.style={postproc cell content/.style={@cell content={\textbf{##1}}}}, 
    every row 1 column 3/.style={postproc cell content/.style={@cell content={\textbf{##1}}}}, 
    every row 1 column 4/.style={postproc cell content/.style={@cell content={\textbf{##1}}}}, 
    every row 1 column 5/.style={postproc cell content/.style={@cell content={\textbf{##1}}}}, 
    every row 1 column 6/.style={postproc cell content/.style={@cell content={\textbf{##1}}}}, 
    every row 1 column 7/.style={postproc cell content/.style={@cell content={\textbf{##1}}}}, 
    every row 1 column 8/.style={postproc cell content/.style={@cell content={\textbf{##1}}}}, 
    every row 1 column 9/.style={postproc cell content/.style={@cell content={\textbf{##1}}}}, 
    every row 1 column 10/.style={postproc cell content/.append style={@cell content={\textbf{##1}}}, zerofill, precision=2,}, 
    ] {\totalbilltable}
\label{table:peak_shaving_demand}
\end{table*}

\noindent \textbf{Heuristics:}~\Cref{alg:charge_first_llf} and~\Cref{alg:llf} show the exact algorithms used for Charge First LLF and Trickle LLF, respectively.

\noindent \textbf{Peak Shaving Results:}~\Cref{table:peak_shaving_demand} shows the peak shaving performance of all approaches. 

\noindent \textbf{Ablation results:} Finally,~\Cref{table:ablation_study} shows a breakdown of the ablation results per month.

\end{document}